\documentclass{article}

\PassOptionsToPackage{numbers,sort&compress}{natbib}
\usepackage[preprint]{neurips_2026}

\PassOptionsToPackage{numbers, sort&compress}{natbib}

\usepackage[utf8]{inputenc}
\usepackage[T1]{fontenc}
\usepackage{amsmath,amssymb,amsfonts}
\usepackage{graphicx}
\usepackage{booktabs}
\usepackage{multirow}
\usepackage{microtype}
\usepackage{xcolor}
\usepackage{url}
\usepackage{natbib}
\usepackage{enumitem}
\usepackage{listings}
\usepackage{makecell}
\usepackage{float}
\usepackage{xr}

\usepackage[hidelinks]{hyperref}
\usepackage[capitalize,nameinlink]{cleveref}

\setlist{nosep,leftmargin=*}
\newcommand{\E}{\mathbb{E}}
\newcommand{\Prob}{\mathbb{P}}
\newcommand{\vo}{\overline{VO}}
\newcommand{\Cov}{\operatorname{Cov}}

\title{Beyond Accuracy: Decomposing the Reasoning Efficiency of LLMs}

\author{
    Daniel Kaiser \thanks{Corresponding author}~~\thanks{Integreat - Norwegian Centre for knowledge-driven machine learning}~~\thanks{UiT - The Arctic University of Norway}
    \And Arnoldo Frigessi\footnotemark[1]~~\thanks{University of Oslo}
    \And Ali Ramezani-Kebrya\footnotemark[1]~~\footnotemark[3]
    \And Benjamin Ricaud\footnotemark[1]~~\footnotemark[2]\And
    \texttt{m@dk.fo} 
}

\date{}

\begin{document}
\maketitle

\begin{abstract}
As reasoning LLMs increasingly trade tokens for accuracy through deliberation, search, and self-correction, a single accuracy score can no longer tell whether those tokens buy useful reasoning, recovery from hard instances, or unnecessary verbosity. We introduce a \emph{trace-optional evaluation protocol} that exactly decomposes token efficiency using three observables available even for closed models: completion rate, conditional correctness given completion, and generated length. When instance-level workload metadata is available, we further normalize generated length by declared task-implied work and separate mean verbalization overhead from workload-dependent scaling. When such metadata is absent, we define an auditable solver-derived workload scale and evaluate its stability under leave-self-out, leave-top-k, and held-out-reference-pool perturbations. We evaluate 14 shared open-weight models on CogniLoad, GSM8K, ProofWriter, and ZebraLogic. We further evaluate 11 additional models on CogniLoad, enabling a fine-grained analysis of reasoning-task difficulty factors: task length, intrinsic difficulty, and distractor density. Efficiency and overhead rankings remain stable across all benchmark pairs, more robustly than accuracy rankings, while the decomposition separates \emph{logic-limited}, \emph{context-limited (truncation-driven)}, and \emph{verbosity-limited} failure modes that look identical under accuracy-per-token.
\end{abstract}

\section{Introduction}
\label{sec:intro}

Accuracy alone is no longer sufficient for reasoning evaluation. Contemporary large language models (LLMs) can improve answers by spending more inference-time tokens through longer chains of thought, search, deliberation, or self-verification \citep{snell2024scaling,muennighoff2025s1,guo2025deepseek}. A model that solves a problem after 500 generated tokens and one that solves it after 10,000 tokens therefore have different operational behavior even when their accuracy is identical.

Reporting accuracy per generated tokens is a useful first correction attempt, but it is still not diagnostic as it collapses distinct failure modes. Low efficiency may mean that a model reaches the output budget without producing a valid answer, completes but answers incorrectly, or answers correctly while generating far more text than the task requires. These cases call for different follow-up tests: context management for truncation, reasoning interventions for conditional errors, and length control or adaptive stopping for verbosity.

We turn token efficiency from a scalar ranking into a diagnostic reporting protocol. The contribution is not another leaderboard score, but an accounting standard: using correctness labels, truncation status, and generated-token counts already produced by standard evaluations for both open models and closed APIs, the protocol identifies which measured component explains a model's inefficiency. The workload layer adds instance-level task-implied work, when available, to distinguish average verbalization overhead from workload-dependent scaling. If no native workload metadata exists for a benchmark, a solver-derived workload supplies an annotation-free fallback under an explicitly declared reference pool. Visible traces, when available, add deterministic compression diagnostics for redundancy and information density, but the {\it core protocol requires no trace access, hidden states, human process labels, or LLM judges}.

Our empirical design has two parts. CogniLoad is a controlled calibration benchmark because it varies task length $N$, intrinsic difficulty $d$, and distractor fraction $\eta$ independently within one state-tracking and deductive-reasoning task family. GSM8K, ProofWriter, and ZebraLogic further validate whether the same efficiency quantities behave coherently across arithmetic, proof reasoning, and constraint satisfaction. Across the 14 open-weight models shared by all four benchmarks, pairwise Spearman correlations of our efficiency measure $E_0$ range from 0.42 to 0.88 and mean workload-normalized overhead correlations range from 0.43 to 0.79. The median accuracy correlation (0.343) is roughly half as high as the median efficiency correlation (0.644), but this pattern is not uniform and has wider bootstrap intervals (\cref{sec:transfer}; see \cref{app:cross-benchmark-uncertainty} for intervals). The decomposition also shows that many low-efficiency models are logic-limited rather than merely verbose, while a separate group is context-limited by truncation.

\textbf{Contributions.}
We contribute: (i) a trace-optional, provider-agnostic \emph{measurement and reporting protocol} that {\it standardizes the decomposition of generated-token efficiency for open and closed models}; (ii) native and solver-derived workload scales with leave-self-out, leave-top-$k$, and held-out-pool audits for the solver-derived fallback; (iii) a cross-benchmark study over CogniLoad, GSM8K, ProofWriter, and ZebraLogic showing that {\it efficiency and mean-overhead rankings usually transfer more strongly than accuracy rankings}; (iv) diagnostic evidence separating logic-, context-, verbosity-, and redundancy-limited regimes; and (v) a public reporting template and evaluation artifact implementing the protocol.

\section{Related work}
\label{sec:related}

\textbf{Inference-time scaling and overthinking.} Inference-time scaling allocates more test-time compute to improve reasoning \citep{snell2024scaling,muennighoff2025s1,guo2025deepseek}. It also makes inefficient reasoning visible: long traces can contain useful intermediate work, but also hesitation loops, restatements, and error-amplifying self-doubt \citep{yang2025towards,chen2024overthinking,sui2025stopoverthinkingsurveyefficient,rcp2025}. We measure which component of a model's inference-time behavior accounts for observed token efficiency, rather than proposing a new decoding strategy.

\textbf{Efficiency-aware reasoning evaluation.} Budget-aware metrics, efficiency--accuracy tradeoffs, and length-control methods evaluate quality jointly with cost \citep{wang2024reasoningtokeneconomiesbudgetaware,du2025ockbench,ma2025cotvalvelengthcompressiblechainofthoughttuning}. Our contribution is complementary: instead of optimizing a budget or fitting a single Pareto frontier, we decompose an observed efficiency score into truncation, conditional incorrectness, mean overhead, and workload-dependent overhead.

\textbf{Process, trace, and workload evaluation.}
Step-level and process-supervision methods can evaluate intermediate reasoning, but often require human labels, reference traces, reward models, or LLM judges \citep{lightman2023lets,evaluatingstepbystep2025}. Our core protocol avoids that requirement and works when traces are hidden. When traces are visible, our optional diagnostics measure surface redundancy without claiming semantic trace correctness. We also use benchmark workload structure where available: CogniLoad exposes generator structure, ProofWriter proof depth, GSM8K gold-solution structure, and ZebraLogic constraint structure \citep{cogniload,cobbe2021training,tafjord2021proofwriter,zebralogic2025}. Solver-derived workload extends this idea to benchmarks without native metadata for a workload measure.

\section{Evaluation protocol}
\label{sec:protocol}

Our protocol is layered: each layer adds one data requirement and one form of diagnostic resolution. {\it The outcome layer is the minimum report and applies to any model and any task}. The workload layer improves the diagnostics using instance-level workload measures. The trace layer is optional and applies only when the reasoning trace is available.

\begin{table}[t]
\caption{Layered reporting protocol.}
\label{tab:protocol-layers}
\centering
\small
\begin{tabular}{p{0.08\linewidth}p{0.30\linewidth}p{0.50\linewidth}}
\toprule
Layer & Required data & Main diagnostic question \\
\midrule
Outcome & correctness $S$, generated tokens $T$, truncation flag $C$ & Did inefficiency come from not finishing, finishing incorrectly, or excessively long reasoning? \\
Workload & outcome data plus instance workload $W$ & How many tokens does the model spend per unit of task-implied work? \\
Trace & workload data plus visible trace text & Is long visible reasoning redundant or relatively information-dense after compression? \\
\bottomrule
\end{tabular}
\end{table}

\subsection{Outcome layer: Token efficiency}
\label{sec:outcome-layer}

Let $\mathcal{D}=\{I_1,\ldots,I_n\}$ be a finite benchmark consisting of benchmark instances $I_i$. For each instance, a fixed model produces an answer with a correctness indicator $S_i\in\{0,1\}$ and generated-token count $T_i>0$. Probabilities and expectations are empirical over instances. Our primary statistic is correct answers per 1,000 generated output tokens:
\begin{equation}
E_0 := \frac{1000\,\Prob(S=1)}{\E[T]}.
\label{eq:e0}
\end{equation}
Equivalently, $\widehat{E}_0=1000\sum_i S_i/\sum_i T_i$ is the pooled throughput under a shared output-token budget. We do not average $S_i/T_i$ over instances because that alternative can overweight short lucky successes.

Let $C$ denote budget truncation and $N=C^c$ clean completion. Define $r_{\mathrm{ctx}}:=\Prob(N)$ as the clean-completion rate, and $r_{\mathrm{logic}}:=\Prob(S=1\mid N)$ as conditional correctness given clean completion. A successful run must first complete, so
\begin{equation}
E_0=1000\cdot\frac{r_{\mathrm{ctx}}r_{\mathrm{logic}}}{\E[T]}.
\label{eq:e0-factor}
\end{equation}
For two models evaluated on the same instance set, log differences decompose exactly:
\begin{equation}
\Delta\log E_0=\Delta\log r_{\mathrm{ctx}}+
\Delta\log r_{\mathrm{logic}}-\Delta\log \E[T].
\label{eq:outcome-log}
\end{equation}
The terms above {\it distinguish budget failure, conditional reasoning failure, and mean generated length.}

\subsection{Workload layer: verbalization relative to task-implied work}
\label{sec:workload-layer}

Raw length is ambiguous because instances differ in how much work they imply.  A long proof and a one-step arithmetic question should not be normalized in the same way.  We therefore attach each instance to a positive workload scale $W(I)>0$.  $W$ is an external normalization, not a model-internal compute measure.

Given $W$, define verbalization overhead $VO=T/W$. Since $T=W\cdot VO$ point-wise, we have 
\begin{equation}
\E[T]=\E[W]\,\vo\,\kappa,
\qquad \vo:=\E[VO],
\label{eq:workload-factor}
\end{equation}
where
\begin{equation}
\kappa:=\frac{\E[W\cdot VO]}{\E[W]\E[VO]}
=1+\frac{\Cov(W,VO)}{\E[W]\E[VO]}.
\label{eq:kappa}
\end{equation}

Mean overhead $\overline{VO}$ is generated tokens per workload unit. Coupling $\kappa$ asks whether overhead grows with workload: $\kappa>1$ indicates higher per-work cost on harder instances, $\kappa\approx1$ workload-independent overhead, and $\kappa<1$ sublinear scaling. Because $\kappa$ depends on workload geometry, we use it mainly for within-benchmark diagnosis rather than as a standalone transferable model ranking.

When all models are evaluated on the same instance set, $\E[W]$ is shared.  Substituting \Cref{eq:workload-factor} into \Cref{eq:e0-factor} yields
\begin{equation}
\Delta\log E_0
=\Delta\log r_{\mathrm{ctx}}+
\Delta\log r_{\mathrm{logic}}-
\Delta\log\vo-
\Delta\log\kappa.
\label{eq:workload-log}
\end{equation}

\paragraph{Native workload.}

Native workload $W_{\mathrm{nat}}$ uses benchmark structure. CogniLoad uses the generator-derived person-of-interest workload in \cref{eq:wpoi}; GSM8K uses gold-solution step count; ProofWriter uses proof depth; and ZebraLogic counts parsed structural constraint atoms, with alternative structural scales in the appendix (\cref{app:workload-scales}). Native workload is preferred for semantic interpretation because its units are tied to the benchmark.

\paragraph{Solver-derived workload.}

Solver-derived workload $W_{\mathrm{solv}}$ is an annotation-free fallback that measures verbosity relative to the most token-efficient successful reference model on each instance. Given a benchmark and reference pool, rank reference models by benchmark-level $E_0$ and set $W_{\mathrm{solv}}(I)$ to the generated length of the highest-ranked reference model that answers $I$ correctly, falling through to the next correct model as needed. The scale is operational and reference-pool-dependent, not an oracle proof length or benchmark-independent ground truth. We use it only with audits: it recovers native mean-overhead rankings where native workload exists (\cref{tab:solver-calibration}) and remains stable under leave-self-out, leave-top-$k$, and held-out-pool perturbations (\cref{tab:wsolv-stress}).

\subsection{Trace layer: auxiliary redundancy diagnostics}
\label{sec:trace-layer}

Motivated by the empirical observation that weaker models often reach the context limit with redundant visible reasoning, we extend the efficiency decomposition for trace-visible models using a benchmark-agnostic compression signal. We compute this signal as the bounded zlib compression ratio of the trace text; see~\cref{app:trace-impl} for implementation details. For trace text, define
\[
\sigma_c(I)=\min\left(1,\frac{|\mathrm{zlib}(\mathrm{trace}(I))|}{|\mathrm{trace}(I)|}\right),
\]
where lower values indicate more compressible and therefore more redundant text.  We define compressed-signal tokens $T_{\mathrm{sig}}(I)=T(I)\sigma_c(I)$ and a trace-quality density factor
\begin{equation}
q_{\mathrm{trace}}:=\frac{\E[T_{\mathrm{sig}}]}{\E[T]}.
\end{equation}
The same workload decomposition can be applied to $T_{\mathrm{sig}}$. This layer is deliberately auxiliary and lightweight: it can distinguish redundant long traces from denser long traces, but it does \emph{not} verify step-by-step semantic correctness, task relevance, or proof validity.

\section{Experimental design}
\label{sec:experimental-design}

\subsection{Benchmarks and workload scales}
\label{sec:benchmarks}

CogniLoad is the controlled calibration benchmark in our study. Each instance gives an initial state for several people, $N$ sequential conditional updates that update it, and a query about the final state of a person of interest (PoI). Task length $N$, intrinsic difficulty of updates $d$, and needle fraction $\eta$ vary independently enabling analysis along controlled difficulty surfaces. Its native workload is

\begin{equation}
\begin{aligned}
W_{\mathrm{poi}}(I)
&= \sum_{t=1}^{N} k_t
 + \sum_{t\in\mathcal{N}(I)} m_t .
\end{aligned}
\label{eq:wpoi}
\end{equation}
where $k_t$ is the number of condition clauses in statement $t$, $m_t$ is the number of update clauses, and $\mathcal{N}(I)$ contains statements affecting the PoI. This corresponds to an oracle that tracks only the queried person while still checking each statement's conditions.

GSM8K, ProofWriter, and ZebraLogic are transfer benchmarks: arithmetic word problems, deductive proof reasoning, and constraint satisfaction. Their native workload scales are coarser than CogniLoad's generator workload, so we treat them as benchmark-grounded proxies. GSM8K uses gold-solution step count, with arithmetic-operation count as an auxiliary scale. ProofWriter uses proof depth. ZebraLogic counts parsed structural clue atoms, with weighted and grid-augmented variants as sensitivity checks.

\subsection{Models, prompting, and sample sizes}
\label{sec:models-protocol}

On CogniLoad we evaluate 25 models, including proprietary API systems and open-weight reasoning models. Cost-constrained API systems use a fixed subset of about 1,400 instances per model, or 10 instances per CogniLoad parameter configuration. Locally served or lower-cost models use about 14,000 instances per model, or 100 per configuration. The smaller subset is nested in the larger suite and has nearly identical mean workload, reducing concern that comparisons reflect configuration mix. Cross-benchmark transfer uses the 14 open-weight models shared across CogniLoad, GSM8K, ProofWriter, and ZebraLogic. Trace diagnostics use the 12 CogniLoad models with visible reasoning traces, totaling 142,800 traces, plus trace-visible local runs on the transfer benchmarks.

All runs use benchmark-default prompting, deterministic answer extraction, provider-recommended decoding settings, and a 32,768-token combined input--output budget chosen to fit the smallest context window in the suite. The primary cost metric $T$ is generated output tokens, including reasoning and final answer. APIs report generated-token counts even when internal reasoning traces are hidden, making this metric available across open and closed systems. Generated tokens are not total compute: they do not capture FLOPs, latency, energy, hidden reasoning tokens, or all implementation-specific costs. We therefore report E0 as visible-output efficiency and treat GPU-hours, API spending, elapsed time, and total-token counts as parallel resource audits (\cref{app:robustness} and \cref{app:resources}).

\section{Results: from scalar efficiency to diagnostic regimes}
\label{sec:results}

\subsection{Efficiency is not accuracy}
\label{sec:accuracy-efficiency}

\begin{figure}[t]
\centering
\includegraphics[width=0.60\linewidth]{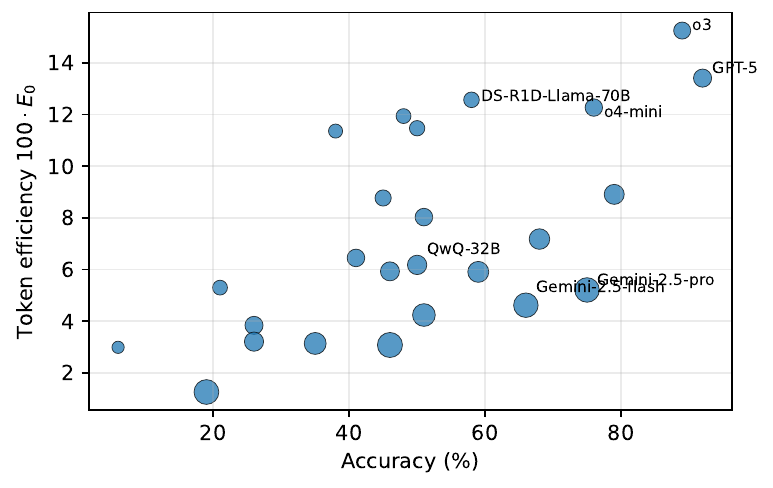}
\caption{Accuracy versus token efficiency on CogniLoad.  Marker area scales with mean generated tokens.  Similar-accuracy models can differ sharply in token cost, and higher accuracy need not imply higher efficiency.}
\label{fig:acc-e0}
\end{figure}

On CogniLoad, accuracy and efficiency rankings disagree (Spearman $\rho=0.63$). \Cref{fig:acc-e0} shows the divergence. GPT-5 is more accurate than o3 (0.92 vs. 0.89), but generates more tokens on average (6,897 vs. 5,869), so its token efficiency is lower. o4-mini and Gemini~2.5~Pro have similar accuracy (0.76 vs. 0.75), but o4-mini uses far fewer tokens (6,232 vs. 14,416), producing a roughly $2.3\times$ efficiency gap. Accuracy remains essential, but it is not a sufficient statistic once models can vary inference effort.

\subsection{Similar efficiency deficits have different causes}
\label{sec:decomposition-results}

\begin{figure}[t]
\centering
\includegraphics[width=0.98\linewidth]{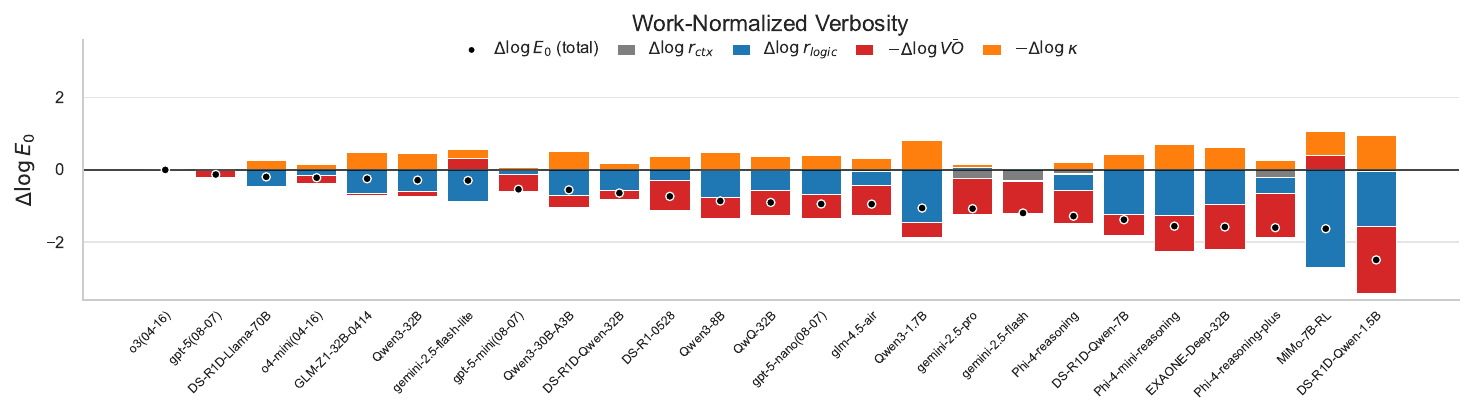}
\caption{Workload-normalized efficiency decomposition on CogniLoad relative to o3.  Bars decompose $\Delta\log E_0$ into completion rate, conditional correctness, mean overhead, and workload coupling.  Many lower-efficiency models are logic-limited, while a separate group is context-limited by truncation.}
\label{fig:decomp}
\end{figure}

The decomposition turns an efficiency gap into a bottleneck profile. \Cref{fig:decomp} applies the workload-normalized decomposition on CogniLoad relative to o3. For many models, the dominant negative term is $r_{\mathrm{logic}}$: they complete and are not primarily verbose relative to frontier systems, but are less likely to be correct conditional on completion. These models are logic-limited. Gemini 2.5 Pro and Gemini 2.5 Flash show a different profile: they have high conditional correctness when complete, but frequent budget truncation, making $r_{\mathrm{ctx}}$ the dominant efficiency loss. Native mean overhead also varies from 13.7 to 127.0 tokens per workload unit, a roughly nine-fold spread. Most cross-model verbosity differences are explained by mean overhead $\overline{VO}$ rather than coupling $\kappa$.

\subsection{Solver-derived workload recovers native mean-overhead rankings}
\label{sec:solver-calibration}

Solver-derived workload is useful only if it preserves the comparative information carried by native workload when native workload exists. \Cref{tab:solver-calibration} shows that $\vo_{\mathrm{solv}}$ closely recovers $\vo_{\mathrm{nat}}$ rankings across all benchmarks. This is the main intended use of $W_{\mathrm{solv}}$: an auditable fallback for comparing average workload-normalized verbosity when native metadata is unavailable.

\begin{table}[t]
\caption{Calibration of solver-derived workload against native workload. The main claim is for mean overhead. For GSM8K, $W_{\mathrm{step}}$ is the primary native scale and $W_{\mathrm{op}}$ is an auxiliary arithmetic-operation scale.}
\label{tab:solver-calibration}
\centering
\small
\begin{tabular}{lccc}
\toprule
Benchmark / native scale & Unique $W$ values & $\rho(\vo_{\mathrm{nat}},\vo_{\mathrm{solv}})$ & $\rho(\kappa_{\mathrm{nat}},\kappa_{\mathrm{solv}})$ \\
\midrule
CogniLoad $W_{\mathrm{poi}}$ & -- & 0.982 & 0.933 \\
GSM8K $W_{\mathrm{step}}$ & 9 & 0.938 & 0.358 \\
GSM8K $W_{\mathrm{op}}$ & 17 & 0.965 & 0.644 \\
ProofWriter $W_{\mathrm{depth}}$ & 6 & 1.000 & 0.925 \\
ZebraLogic $W_{\mathrm{atom}}$ & 29 & 0.969 & 0.934 \\
\bottomrule
\end{tabular}
\end{table}

This result also clarifies the division of labor between workload scales. Native workload is preferred when the goal is semantic interpretation because its units are tied to the benchmark abstraction.  Solver-derived workload is a practical fallback for mean-overhead comparison when native metadata is absent.  On GSM8K, counting arithmetic operations in the gold solution gives a finer workload scale than gold-step count and improves both mean-overhead and coupling agreement. On ProofWriter, the perfect mean-overhead agreement is not a tied-rank artifact: Kendall's $\tau_b$ is also 1.000 and the maximum rank difference is zero. On ZebraLogic, mean-overhead rankings are invariant across the tested structural workload variants, including weighted clue-type and grid-augmented definitions. Thus the solver-derived scale is not a substitute for semantic workload metadata, but it is a robust fallback for ranking average workload-normalized verbosity.

Because $W_{\mathrm{solv}}$ is reference-pool-dependent, we audit it under leave-self-out, leave-top-$k$, and held-out-reference-pool perturbations. \Cref{tab:wsolv-stress} shows that mean-overhead rankings remain stable, although CogniLoad is the most sensitive benchmark. The top solver often defines most solved workloads, so reports should release the reference ordering and per-instance solver identity. We treat $W_{\mathrm{solv}}$ as an auditable operational scale for mean overhead, not as a benchmark-independent estimate of irreducible reasoning work.

\begin{table}[t]
\caption{Reference-pool stress tests for solver-derived mean overhead. Leave-top reports the weakest $\rho(\vo)$ over leave-top-1, -3, and -5. Held-out reports the median and 5--95\% interval over 200 random reference/evaluation splits.}
\label{tab:wsolv-stress}
\centering
\small
\begin{tabular}{lcccc}
\toprule
Benchmark & Leave-self $\rho(\vo)$ & Worst leave-top $\rho(\vo)$ & Held-out median $\rho(\vo)$ & Top-solver share \\
\midrule
CogniLoad   & 0.941 & 0.772 & 0.841 [0.637, 0.995] & 0.904 \\
GSM8K       & 1.000 & 0.982 & 1.000 [0.964, 1.000] & 0.975 \\
ProofWriter & 0.996 & 0.973 & 1.000 [0.998, 1.000] & 0.692 \\
ZebraLogic  & 0.974 & 0.995 & 1.000 [0.893, 1.000] & 0.911 \\
\bottomrule
\end{tabular}
\end{table}

These audits define the scope of the fallback: it is suitable for auditable mean-overhead comparisons, but not for estimating irreducible reasoning work.

\subsection{Efficiency and mean overhead transfer across benchmarks}
\label{sec:transfer}

A useful protocol should not only explain one benchmark. \Cref{tab:pairwise-transfer} reports pairwise Spearman correlations over the 14 open-weight models shared by CogniLoad, GSM8K, ProofWriter, and ZebraLogic. Accuracy correlations range from 0.288 to 0.777, with median 0.343. Efficiency correlations are positive for every pair, with $\rho(E_0)$ from 0.424 to 0.881 and median 0.644. Mean workload-normalized overhead is also consistently positive, with $\rho(\vo_{\mathrm{solv}})$ from 0.429 to 0.785 and median 0.732. $E_0$ and $\vo_{\mathrm{solv}}$ exceed accuracy in five of six pairs; the exception is GSM8K--ZebraLogic, where accuracy itself transfers strongly. These correlations do not make one benchmark a substitute for another, but they show that generated-token efficiency and mean overhead capture model-level tendencies that accuracy often misses.

\begin{table}[t]
\caption{Cross-benchmark rank correlations over the 14 shared open-weight models. $\vo_{\mathrm{solv}}$ and $\kappa_{\mathrm{solv}}$ use solver-derived workload so that the workload layer is defined uniformly on every benchmark. Bootstrap intervals are reported in \Cref{app:cross-benchmark-uncertainty}.}
\label{tab:pairwise-transfer}
\centering
\small
\setlength{\tabcolsep}{3.5pt}
\begin{tabular}{lcccc}
\toprule
Benchmark pair & $\rho(E_0)$ & $\rho(\mathrm{Acc})$ & $\rho(\vo_{\mathrm{solv}})$ & $\rho(\kappa_{\mathrm{solv}})$ \\
\midrule
CogniLoad--GSM8K       & 0.881 & 0.587 & 0.705 & $-$0.354 \\
CogniLoad--ProofWriter & 0.705 & 0.332 & 0.758 & $-$0.275 \\
CogniLoad--ZebraLogic  & 0.499 & 0.354 & 0.771 & $-$0.231 \\
GSM8K--ProofWriter     & 0.780 & 0.304 & 0.785 & 0.807 \\
GSM8K--ZebraLogic      & 0.424 & 0.777 & 0.604 & 0.411 \\
ProofWriter--ZebraLogic& 0.582 & 0.288 & 0.429 & 0.516 \\
\bottomrule
\end{tabular}
\end{table}

The asymmetry between $\vo$ and $\kappa$ is important. Mean overhead is a broad model-level tendency: some models spend more generated text per unit of declared work across tasks. Coupling is a within-benchmark shape statistic: it asks whether overhead grows or shrinks with that benchmark's workload geometry. Bootstrap intervals for $\kappa$ often cross zero (\cref{app:cross-benchmark-uncertainty}), so we use $\vo$ for cross-benchmark comparison and $\kappa$ for local diagnosis.

\subsection{Long sequential contexts dominate CogniLoad efficiency loss}
\label{sec:difficulty-results}

\begin{figure}[t]
\centering
\includegraphics[width=0.98\linewidth]{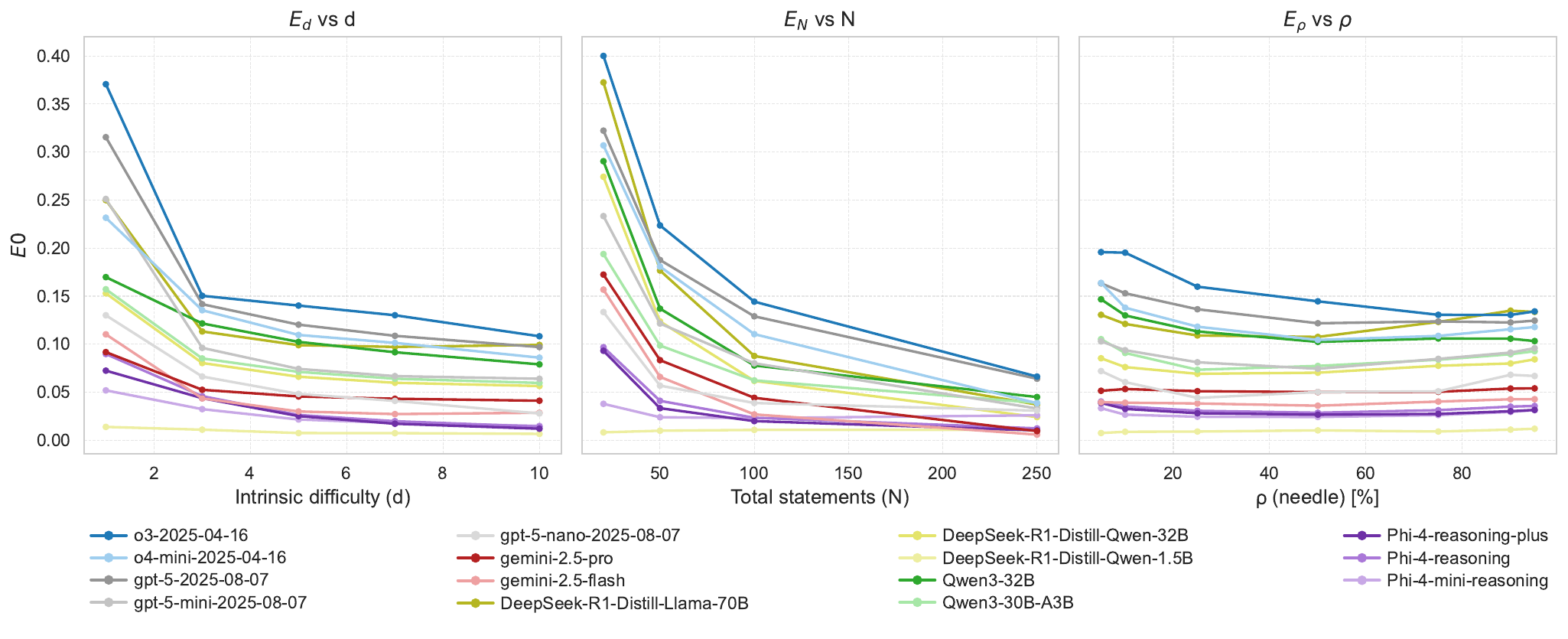}
\caption{Efficiency across CogniLoad difficulty surfaces.  Task length $N$ produces the largest efficiency loss; intrinsic difficulty $d$ has a strong early effect and diminishing marginal impact; needle fraction $\eta$ produces smaller and often U-shaped changes.}
\label{fig:difficulty}
\end{figure}

CogniLoad's factorial design identifies which controlled difficulty axes most affect visible-token efficiency.  \Cref{fig:difficulty} shows that task length is the dominant axis: increasing N from 20 to 250 reduces $E_0$ by roughly 70--90\% across most models. Intrinsic difficulty also reduces efficiency, but much of the loss occurs when moving from easy to non-trivial reasoning and then tapers. Needle fraction $\eta$ has a weaker and often U-shaped effect, consistent with a tradeoff between sparse relevance detection and a high volume of relevant updates.

This result gives the cross-benchmark findings a mechanistic anchor. The transfer analysis shows that efficiency is not a CogniLoad-only artifact; CogniLoad then shows where efficiency degrades under controlled stress. In this suite, long sequential context is a stronger efficiency stressor than local intrinsic difficulty or distractor density.

\subsection{Optional compression diagnostics separate redundant from dense verbosity}
\label{sec:trace-results}

\begin{figure}[t]
\centering
\includegraphics[width=0.98\linewidth]{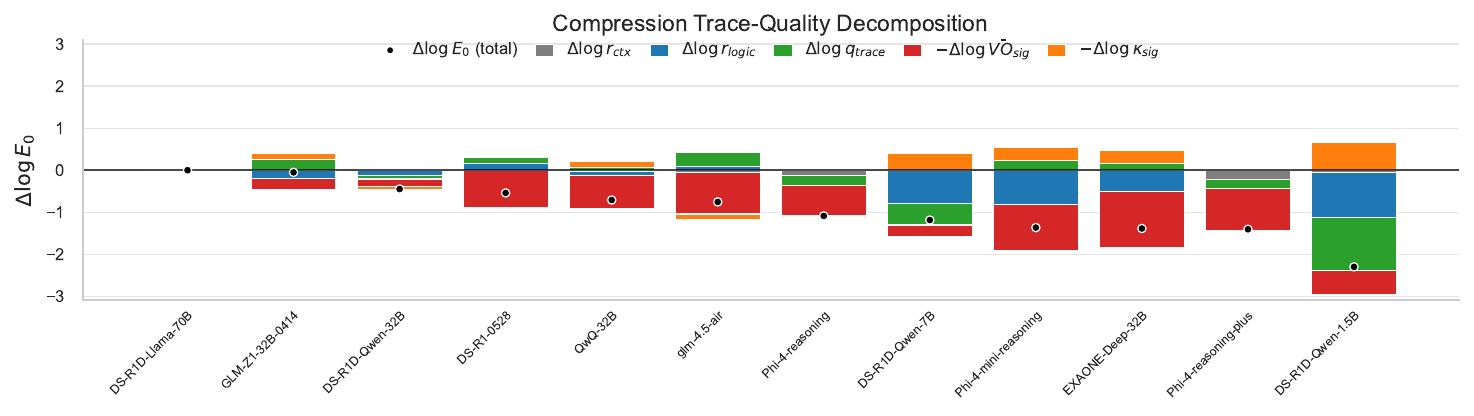}

\caption{Compression-trace decomposition for the 12 trace-visible CogniLoad models, using DeepSeek-R1-Distill-Llama-70B as the reference. The trace layer separates redundancy-dominated collapse from high compressed-signal overhead; it is not a semantic proof checker.}
\label{fig:trace-main}
\end{figure}

When visible traces are available, the trace layer separates two forms of long generation that raw token counts conflate. \Cref{fig:trace-main} shows that compression density varies by about a factor of five ($q_{\mathrm{trace}}\in[0.052,0.255]$) across the 12 trace-visible CogniLoad models. DeepSeek-R1-Distill-Qwen-1.5B produces especially compressible visible text, consistent with redundant repetition. EXAONE-Deep-32B has moderate compression density but high compressed-signal overhead ($\bar{VO}_{\mathrm{sig}}=15.35$), indicating dense but costly verbosity. Phi-4-reasoning and Phi-4-reasoning-plus combine low compression density with high compressed-signal overhead, consistent with redundant long reasoning.


These diagnostics remain auxiliary. Compression detects repeated surface form; it does not certify semantic correctness or task-relevant reasoning. The core protocol and all cross-benchmark claims rely on final-answer correctness, truncation status, and token counts. The truncation audit provides a separate check on context-limited failures: truncation is concentrated in a small number of model/benchmark pairs, with DeepSeek-R1-Distill-Qwen-1.5B on ZebraLogic as the most extreme case with an 85.9\% truncation. On CogniLoad, several local models have truncated tails with high repeated 4-gram overlap, supporting the interpretation that some context failures are budget exhaustion through redundant generation rather than ordinary long reasoning.

\subsection{Robustness checks}
\label{sec:robustness-results}

The visible-token rankings are not driven by tokenizer artifacts or the API/local sample-size difference. Character-normalized rankings correlate strongly with token-normalized rankings \(\rho=0.98\), and replacing output tokens with total tokens leaves transfer-benchmark rankings nearly unchanged \(\rho \in [0.987, 1.000]\). Subsampling local-model runs to the API-cost-constrained 10-instance-per-configuration regime recovers full-run estimates well, with median relative error 2.0\% for \(E_0\) and 1.2\% for \(\overline{VO}\) and rank \(\rho \geq 0.995\). Deployment-cost views can differ: elapsed-time and GPU-hour rankings are not equivalent to visible-token efficiency, so we treat them as parallel resource audits rather than replacements for \(E_0\). Excluding truncations, incorrect completions average 1.23 times more tokens than correct ones, ruling out early abandonment as the sole explanation for sublinear  \(\kappa\). \cref{app:robustness} provides the full audit.

\section{Reporting standard and interpretation}
\label{sec:using-protocol}

A reasoning-efficiency report should include, at minimum, accuracy, $E_0$, mean generated length $\E[T]$, completion rate $r_{\mathrm{ctx}}$, and conditional correctness $r_{\mathrm{logic}}$. These quantities require no trace access and can be computed for closed APIs as well as open models. With native workload metadata, reports should additionally include mean workload-normalized overhead $\overline{VO}_{\mathrm{nat}}$ and coupling $\kappa_{\mathrm{nat}}$, with $\kappa$ interpreted only within the benchmark. Without native metadata, $\overline{VO}_{\mathrm{solv}}$ should be reported only with a declared reference pool, per-instance reference-solver identity, and at least one pool-sensitivity check. When traces are visible, $q_{\mathrm{trace}}$, $\overline{VO}_{\mathrm{sig}}$, and $\kappa_{\mathrm{sig}}$ should be reported as compression-based redundancy and density diagnostics, not as semantic trace-quality measures.

The protocol is an accounting tool, not a causal model of reasoning. It does not infer hidden compute, prove that long traces are unnecessary, or certify the validity of intermediate reasoning. Its purpose is to identify which measured component of an evaluation run accounts for visible-token inefficiency, so that subsequent interventions can be targeted and tested. \Cref{tab:diagnostic-actions} summarizes the resulting diagnostic signatures and follow-up tests; \cref{app:what-transfer} states which quantities should and should not be expected to transfer across benchmarks.

\begin{table}[t]
\caption{Diagnostic signatures and follow-up tests. The protocol identifies which measured component limits visible-token efficiency; it does not by itself prove that a proposed intervention will improve that component.}
\label{tab:diagnostic-actions}
\centering
\small
\begin{tabular}{p{0.18\linewidth}p{0.31\linewidth}p{0.37\linewidth}}
\toprule
Signature & Measured pattern & Follow-up to test \\
\midrule
Logic-limited & low $r_{\mathrm{logic}}$, moderate overhead & verification, reasoning-data, or curriculum interventions \\
Verbosity-limited & high $r_{\mathrm{logic}}$, high $\vo$ & length control, concise distillation, or adaptive stopping \\
Context-limited & high conditional correctness when complete; low completion rate & loop detection, chunked reasoning, or budget-aware decoding \\
Redundancy-limited & low $q_{\mathrm{trace}}$ with long traces or high compressed-signal overhead & loop detection, length control, or stop rules with accuracy monitoring \\
\bottomrule
\end{tabular}
\end{table}

\section{Limitations}
\label{sec:limitations}

\textbf{Visible tokens are not total compute.}
Generated visible tokens are a practical common unit, but they do not directly measure FLOPs, wall-clock latency, energy, monetary cost, or hidden reasoning tokens. Character-normalized and total-token-normalized rankings are close to the visible-token rankings, but elapsed-time and resource-cost rankings can differ substantially. We therefore interpret $E_0$ as visible-output efficiency. Deployment studies should report $E_0$ alongside total tokens, latency, billed reasoning tokens, card-hours, energy, or dollar cost when those quantities are available.

\textbf{Workload scales are proxies.}
Native workload measures are interpretable but benchmark-specific. GSM8K step count and ProofWriter proof depth are coarse integer-valued signals; ZebraLogic constraint counting depends on a deterministic clue parser; and CogniLoad workload reflects a synthetic state-tracking task. Our sensitivity checks reduce concern that our main mean-overhead conclusions depend on a single arbitrary workload definition, but they do not make workload units commensurable across benchmark families.

\textbf{Solver-derived workload is operational, not an oracle.}
Solver-derived workload extends the workload layer to benchmarks without native metadata, but it depends on the declared reference pool. The top reference solver often defines most solved workloads, so users should release the reference ordering and per-instance solver identity. Our leave-self-out, leave-top-\(k\), and held-out-reference-pool analyses show that mean-overhead rankings are stable under substantial perturbations, but they do not make $W_{\mathrm{solv}}$ an estimate of irreducible proof length. $W_{\mathrm{solv}}$ should therefore be used as an auditable fallback for mean-overhead comparison, not as a benchmark-independent task-complexity measure.

\textbf{Output-budget dependence.}
Completion rates and context-limited profiles depend on the chosen output budget. We use a shared 32,768-token combined input--output budget to make comparisons feasible across the smallest context window in the suite, but models with larger native context windows could show different truncation profiles under model-specific budgets. We therefore interpret context-limited diagnoses as properties of the stated evaluation budget rather than intrinsic model properties.

\textbf{Statistical scope.}
The cross-benchmark transfer analysis uses 14 shared open-weight models.  This overlap is enough to reveal broad positive rank structure for $E_0$ and $\vo$, but it leaves wide uncertainty for fine-grained pairwise claims, especially for $\kappa$.  The API systems are evaluated on a smaller nested CogniLoad subset for cost reasons; those results are appropriate for CogniLoad-level comparison but are not used in the cross-benchmark transfer correlations.

\textbf{Trace diagnostics are surface diagnostics.}
The trace layer measures visible-text redundancy through compression.  It does not verify the semantic validity or task relevance of intermediate reasoning.  Outcome-level correctness, workload-normalized overhead, and trace signal should therefore be reported together rather than collapsed into a single claim that a trace is good or bad.

\section{Conclusion}
\label{sec:conclusion}

Reasoning evaluations should distinguish correct answers from efficiently obtained correct answers. We introduced a trace-optional protocol that decomposes visible-token efficiency into completion rate, conditional correctness, and generated length, and that adds workload-normalized overhead when an instance-level workload scale is available. Across CogniLoad, GSM8K, ProofWriter, and ZebraLogic, visible-token efficiency and mean workload-normalized overhead reveal cross-model regularities that accuracy often misses, while $\kappa$ and trace diagnostics are best interpreted as local diagnostic quantities. CogniLoad further identifies long sequential context as the strongest efficiency stressor in the suite.

The main recommendation is simple: reasoning evaluations should not report only whether a model answered correctly, or even only how many correct answers it produced per token. They should report the accounting behind that efficiency: whether the model failed by truncation, by conditional incorrectness after completion, or by excessive generated text relative to declared task-implied work. This turns token efficiency from a scalar leaderboard into a reproducible diagnostic object.

\section{Author Contributions}
D.K. conceived the idea, developed the decomposition and the solver-derived workload, reviewed the literature, ran and implemented the evaluation of LLMs and experiments, and wrote the first version of the manuscript. A.F. and D.K. developed the outcome-level decomposition. B.R. and D.K. developed the workload-normalized and trace-quality decomposition. All authors developed the notion of reasoning efficiency, and edited the manuscript towards the final version.

\section{Acknowledgements and Support}

This work was supported by the Research Council of Norway through its Centres of Excellence scheme, Integreat - Norwegian Centre for knowledge-driven machine learning, project number 332645. 

Ali Ramezani-Kebrya was supported by the Research Council of Norway through FRIPRO Grant under project number 356103, its Centres of Excellence scheme, Integreat - Norwegian Centre for knowledge-driven machine learning under project number 332645, and its Centre for Research-based Innovation funding scheme (Visual Intelligence under grant no. 309439).

We acknowledge NRIS Norway for awarding this project access to the LUMI supercomputer, owned by the EuroHPC Joint Undertaking, hosted by CSC (Finland) and the LUMI consortium through Sigma2, project number nn12027k.

\clearpage
\bibliographystyle{plainnat}
\bibliography{bibliography}

\appendix

\section*{Appendix organization}

\begin{itemize}
    \item \Cref{app:derivations}: algebra for the outcome, workload, and trace decompositions.
    \item \Cref{app:workload-scales}: native and solver-derived workload definitions, including the CogniLoad oracle workload.
    \item \Cref{app:wsolv-reference-sensitivity}: leave-self-out, leave-top-$k$, held-out-reference-pool, and solver-dominance audits for $W_{\mathrm{solv}}$.
    \item \Cref{app:transfer-workload-sensitivity}: GSM8K, ProofWriter, and ZebraLogic native-workload sensitivity.
    \item \Cref{app:cross-benchmark}: cross-benchmark Pearson/Spearman correlations, pairwise workload correlations,  native/solver calibration, and transfer of protocol quantities across benchmarks.
    \item \Cref{app:models-results}: full model list, CogniLoad leaderboard, and workload-normalized verbosity table.
    \item \Cref{app:trace-impl,app:compression}: compression trace-quality implementation and cross-benchmark checks.    
    \item \Cref{app:robustness,app:alternative-costs,app:oracle-early-stop,app:truncation-loop-audit}: tokenization, sampling, alternative cost columns, oracle early-stop, truncation-loop, selection-effect, and bootstrap checks.
    \item \Cref{app:resources}: per-model compute and API-cost accounting.
    \item \Cref{app:difficulty,app:profiles,app:case-studies}: controlled CogniLoad difficulty analysis, bottleneck profiles, and representative trace case studies.
\end{itemize}

\section{Derivations}
\label{app:derivations}

\subsection{Efficiency definition and empirical estimator}

Let $I\sim\mathcal{D}$ denote a benchmark instance.  A model run on $I$ produces a correctness indicator $S\in\{0,1\}$ and an output token count $T>0$.  We define efficiency as correct answers per thousand generated output tokens:
\begin{equation}
E_0=\frac{1000\,\mathbb{P}(S=1)}{\mathbb{E}[T]}=\frac{1000\,\mathbb{E}[S]}{\mathbb{E}[T]}.
\end{equation}
Given $n$ evaluated instances $\{(S_i,T_i)\}_{i=1}^n$, the estimator is the ratio of sample means,
\begin{equation}
\widehat{E}_0=1000\cdot\frac{\frac{1}{n}\sum_i S_i}{\frac{1}{n}\sum_i T_i}=1000\cdot\frac{\sum_i S_i}{\sum_i T_i}.
\end{equation}
This estimator corresponds to the throughput interpretation: how many correct answers are produced by a fixed global token budget.  The alternative $1000\cdot n^{-1}\sum_i S_i/T_i$ gives each instance equal weight after dividing by its own length.  That alternative can overweight short lucky successes and does not correspond to pooled deployment cost.

\subsection{Outcome partition and robustness factorization}

Let $C$ denote token-budget truncation and $N=C^c$ clean completion.  Define the success event $\mathsf{S}=\{S=1\}$ and the logic-failure event $L=N\cap\mathsf{S}^c$.  Under the evaluation protocol, a successful answer must be completed, so
\begin{equation}
\mathbb{P}(\mathsf{S})=\mathbb{P}(\mathsf{S}\cap N)=\mathbb{P}(N)\mathbb{P}(\mathsf{S}\mid N)=r_{\mathrm{ctx}}r_{\mathrm{logic}},
\end{equation}
where $r_{\mathrm{ctx}}=\mathbb{P}(N)$ and $r_{\mathrm{logic}}=\mathbb{P}(\mathsf{S}\mid N)$.  The empirical estimators are
\begin{equation}
\widehat{r}_{\mathrm{ctx}}=1-\frac{1}{n}\sum_i C_i,\qquad
\widehat{r}_{\mathrm{logic}}=\frac{\sum_i S_i}{\sum_i(1-C_i)}.
\end{equation}
Substituting into $E_0$ gives
\begin{equation}
E_0=1000\cdot\frac{r_{\mathrm{ctx}}r_{\mathrm{logic}}}{\mathbb{E}[T]}.
\end{equation}
For a reference model, log differences give
\begin{equation}
\log\frac{E_0}{E_{0,\mathrm{ref}}}
=\log\frac{r_{\mathrm{ctx}}}{r_{\mathrm{ctx,ref}}}
+\log\frac{r_{\mathrm{logic}}}{r_{\mathrm{logic,ref}}}
-\log\frac{\mathbb{E}[T]}{\mathbb{E}[T]_{\mathrm{ref}}}.
\end{equation}

\subsection{Workload-normalized factorization}

Let $W(I)>0$ be a workload scale and define $VO=T/W$.  Since $T=W\,VO$ pointwise,
\begin{equation}
\mathbb{E}[T]=\mathbb{E}[W\,VO]=\mathbb{E}[W]\mathbb{E}[VO]+\mathrm{Cov}(W,VO).
\end{equation}
Define
\begin{equation}
\bar{VO}:=\mathbb{E}[VO],\qquad
\kappa:=\frac{\mathbb{E}[W\,VO]}{\mathbb{E}[W]\mathbb{E}[VO]}
=1+\frac{\mathrm{Cov}(W,VO)}{\mathbb{E}[W]\mathbb{E}[VO]}.
\end{equation}
Then
\begin{equation}
\mathbb{E}[T]=\mathbb{E}[W]\bar{VO}\kappa.
\end{equation}
When all models are evaluated on the same instance set, $\mathbb{E}[W]$ is shared and cancels in cross-model comparisons:
\begin{equation}
\Delta\log E_0=\Delta\log r_{\mathrm{ctx}}+\Delta\log r_{\mathrm{logic}}-\Delta\log\bar{VO}-\Delta\log\kappa.
\end{equation}

\paragraph{Interpretation of $\kappa$.}
$\kappa>1$ means per-work overhead is positively coupled to workload: the model spends more tokens per unit of work on larger instances.  $\kappa\approx1$ means overhead is approximately workload-independent.  $\kappa<1$ means per-work overhead falls on larger instances.  The latter does not automatically imply a desirable behavior; it can reflect useful compression or selection effects.  \Cref{app:selection} checks this issue empirically.

\subsection{Trace-quality factorization}

For trace-visible models, let $\sigma_c(I)\in[0,1]$ be the bounded zlib compression ratio of the visible trace:
\[
\sigma_c(I)=\min\left(1,\frac{|\mathrm{zlib}(\mathrm{trace}(I))|}{|\mathrm{trace}(I)|}\right).
\]
Let
\begin{equation}
T_{\mathrm{sig}}(I)=T(I)\sigma_c(I),\qquad
q_{\mathrm{trace}}:=\frac{\mathbb{E}[T_{\mathrm{sig}}]}{\mathbb{E}[T]}.
\end{equation}
Define signal overhead $VO_{\mathrm{sig}}=T_{\mathrm{sig}}/W$, mean signal overhead $\bar{VO}_{\mathrm{sig}}=\mathbb{E}[VO_{\mathrm{sig}}]$, and signal coupling
\begin{equation}
\kappa_{\mathrm{sig}}=\frac{\mathbb{E}[W\,VO_{\mathrm{sig}}]}{\mathbb{E}[W]\mathbb{E}[VO_{\mathrm{sig}}]}.
\end{equation}
Because $\mathbb{E}[T]=\mathbb{E}[T_{\mathrm{sig}}]/q_{\mathrm{trace}}$, the relative trace decomposition is
\begin{equation}
\Delta\log E_0=\Delta\log r_{\mathrm{ctx}}+\Delta\log r_{\mathrm{logic}}+\Delta\log q_{\mathrm{trace}}-\Delta\log\bar{VO}_{\mathrm{sig}}-\Delta\log\kappa_{\mathrm{sig}}.
\end{equation}
The trace layer is diagnostic rather than semantic: a high $q_{\mathrm{trace}}$ indicates less compressible visible text, not a formally valid proof or task-grounded explanation.

\section{Workload scales}
\label{app:workload-scales}

\subsection{Native workload definitions}

Native workload uses benchmark-specific structure or gold annotations.  The experiments use the following definitions:
\begin{itemize}
    \item \textbf{CogniLoad:} $W_{\mathrm{poi}}$, the person-of-interest workload in \Cref{eq:wpoi}.  It counts all condition checks and the updates applied to the queried person's state.
    \item \textbf{GSM8K:} $W_{\mathrm{step}}$, the gold-solution step count, used as the primary native scale.  We also compute $W_{\mathrm{op}}$, the number of arithmetic operations in the annotated gold-solution expressions, and $W_{\mathrm{mix}}=W_{\mathrm{step}}+W_{\mathrm{op}}$.
    \item \textbf{ProofWriter:} $W_{\mathrm{depth}}$, the provided proof depth.  This is interpretable but coarse: the balanced split contains six depth values with 500 instances each.
    \item \textbf{ZebraLogic:} $W_{\mathrm{atom}}$, the number of parsed structural clue atoms.  We also compute $W_{\mathrm{weighted}}$, which weights binary exclusion, ordering, distance, and equality clues more heavily than unary position clues, and $W_{\mathrm{grid}}=W_{\mathrm{atom}}+\text{total grid cells}$.
\end{itemize}
The native scale is preferable when the goal is within-benchmark interpretation: $\bar{VO}_{\mathrm{nat}}$ is tokens per semantic or structural unit of work under that benchmark's declared abstraction.  Because these abstractions differ across benchmarks, native workload is used for interpretation and sensitivity analysis rather than for claiming that all workload units are globally commensurable.

\subsection{Solver-derived workload definition}

Solver-derived workload requires no explicit workload annotation.  For each benchmark, rank the reference models by $E_0$ on that benchmark.  For each instance $I$, choose the highest-ranked model that answers $I$ correctly and set $W_{\mathrm{solv}}(I)$ equal to that model's generated-token count on $I$.  If the top-ranked model is incorrect, fall through to the next-ranked correct model.

This definition has three important properties.  First, it is annotation-free: it only needs correctness labels and token counts.  Second, it is benchmark-specific but not benchmark-engineered: the same rule applies to arithmetic, proof, state-tracking, and constraint-satisfaction tasks.  Third, it is evaluation-pool-dependent.  Adding a substantially more concise correct model could change $W_{\mathrm{solv}}$, and including an evaluated model in the reference pool can make the estimate transductive for that model.  We therefore use it as an operational workload scale for comparative analysis, not as a claim about irreducible minimal reasoning length.  Reports using $W_{\mathrm{solv}}$ should state the reference pool and include sensitivity checks when the statistic supports substantive conclusions.

\subsection{Reference-pool sensitivity of solver-derived workload}
\label{app:wsolv-reference-sensitivity}

Solver-derived workload is intentionally operational: it depends on the declared reference pool.  We therefore report three perturbation analyses.

\paragraph{Leave-self-out.}
For each evaluated model, we remove that model from the reference pool used to define its own $W_{\mathrm{solv}}$.  Mean-overhead rankings remain stable in all four benchmarks.

\begin{table}[H]
\centering
\small
\begin{tabular}{lcccc}
\toprule
Benchmark & Solver coverage & $\rho(\vo)$ & $\rho(\kappa)$ & Max $\vo$ rank shift \\
\midrule
CogniLoad   & 0.989 & 0.941 & 0.995 & 9 \\
GSM8K       & 0.990 & 1.000 & 0.864 & 0 \\
ProofWriter & 0.946 & 0.996 & 0.859 & 1 \\
ZebraLogic  & 0.940 & 0.974 & 0.930 & 3 \\
\bottomrule
\end{tabular}
\caption{Leave-self-out sensitivity for solver-derived workload.  CogniLoad uses the 1,325-instance nested subset shared by all 25 CogniLoad models; the transfer benchmarks use their full shared evaluation sets.}
\label{tab:leave-self-out}
\end{table}

\paragraph{Leave-top-$k$.}
We next remove the top one, three, or five reference solvers and recompute $W_{\mathrm{solv}}$.  The table reports each perturbation, including retained instance coverage and rank stability against the original solver-derived workload.

\begin{table}[H]
\centering
\small
\setlength{\tabcolsep}{3.5pt}
\begin{tabular}{llcccc}
\toprule
Benchmark & Removed top solvers & Coverage & Models & $\rho(\vo)$ & $\rho(\kappa)$ \\
\midrule
CogniLoad   & 1 & 0.984 & 24 & 0.908 & 0.797 \\
CogniLoad   & 3 & 0.970 & 22 & 0.772 & 0.244 \\
CogniLoad   & 5 & 0.965 & 20 & 0.878 & 0.138 \\
GSM8K       & 1 & 0.990 & 13 & 0.984 & 0.857 \\
GSM8K       & 3 & 0.990 & 11 & 0.982 & 0.718 \\
GSM8K       & 5 & 0.990 &  9 & 0.983 & 1.000 \\
ProofWriter & 1 & 0.944 & 13 & 1.000 & 0.940 \\
ProofWriter & 3 & 0.941 & 11 & 0.973 & 0.691 \\
ProofWriter & 5 & 0.934 &  9 & 0.983 & 0.650 \\
ZebraLogic  & 1 & 0.936 & 13 & 0.995 & 0.962 \\
ZebraLogic  & 3 & 0.899 & 11 & 1.000 & 0.909 \\
ZebraLogic  & 5 & 0.866 &  9 & 1.000 & 0.983 \\
\bottomrule
\end{tabular}
\caption{Leave-top-$k$ reference-pool perturbations for solver-derived workload.  CogniLoad is the sensitive case; GSM8K, ProofWriter, and ZebraLogic preserve mean-overhead rankings strongly.}
\label{tab:leave-top-k}
\end{table}

\paragraph{Reference-solver dominance.}
The top reference solver often defines most solved workloads, which is why per-instance solver identity must be part of the released artifact.  The dominance is largest on GSM8K and smallest on ProofWriter.

\begin{table}[H]
\centering
\scriptsize
\setlength{\tabcolsep}{3pt}
\begin{tabular}{@{}lrrrrrr@{}}
\toprule
Benchmark
& \makecell{Shared\\instances}
& \makecell{Solver\\coverage}
& \makecell{Top-1\\share}
& \makecell{Top-3\\share}
& \makecell{Effective\\solvers}
& \makecell{No-correct\\reference\\share} \\
\midrule
CogniLoad   & 1325 & 0.989 & 0.904 & 0.976 & 1.570 & 0.011 \\
GSM8K       & 1319 & 0.990 & 0.975 & 0.989 & 1.173 & 0.010 \\
ProofWriter & 3000 & 0.946 & 0.692 & 0.851 & 3.470 & 0.054 \\
ZebraLogic  &  298 & 0.940 & 0.911 & 0.975 & 1.504 & 0.060 \\
\bottomrule
\end{tabular}
\caption{Solver dominance and fallthrough audit for \(W_{\mathrm{solv}}\).}
\label{tab:solver-dominance}
\end{table}


\paragraph{Held-out reference pools.}
Across 200 random half-pool splits, held-out reference pools preserve mean-overhead rankings strongly outside CogniLoad and moderately on CogniLoad.

\begin{table}[H]
\centering
\small
\begin{tabular}{lcccc}
\toprule
Benchmark & Median coverage & Median $\rho(\vo)$ & 5--95\% $\rho(\vo)$ & Median $\rho(\kappa)$ \\
\midrule
CogniLoad   & 0.972 & 0.841 & [0.637, 0.995] & 0.321 \\
GSM8K       & 0.987 & 1.000 & [0.964, 1.000] & 0.857 \\
ProofWriter & 0.902 & 1.000 & [0.998, 1.000] & 0.964 \\
ZebraLogic  & 0.913 & 1.000 & [0.893, 1.000] & 0.964 \\
\bottomrule
\end{tabular}
\caption{Held-out-reference-pool sensitivity for solver-derived workload.}
\label{tab:heldout-reference-pools}
\end{table}

\subsection{CogniLoad oracle workload and necessity of condition checks}
\label{app:oracle}

CogniLoad statements use implicit subject selection.  A statement specifies attribute conditions that determine which people are affected, rather than naming a target person directly.  For example, a statement may refer to ``the people wearing green socks'' and then update their music preference.  To determine whether the person of interest is affected, even an oracle must evaluate the statement's condition clauses against the current state of that person.

The person-of-interest oracle tracks only the queried person $p^*$.  Let $S_t(p^*,c)$ be the value of attribute category $c$ for $p^*$ after processing statements $1,\ldots,t$.  The oracle algorithm is:
\begin{enumerate}
    \item Initialize the tracked state $S\leftarrow S_0(p^*,\cdot)$.
    \item For each statement $t=1,\ldots,N$:
    \begin{enumerate}
        \item check all $k_t$ condition clauses against $S$;
        \item if all conditions are satisfied, apply the $m_t$ updates to $S$.
    \end{enumerate}
    \item Return the queried attribute of $S$.
\end{enumerate}
The algorithm performs exactly
\begin{equation}
W_{\mathrm{poi}}(I)=\sum_{t=1}^N k_t+\sum_{t\in\mathcal{N}(I)}m_t
\end{equation}
operations.  The first term is irreducible for the person-of-interest oracle because the oracle must check every statement to know whether it applies.  The second term counts only updates for needle statements that actually affect the tracked state.

\subsection{Alternative CogniLoad workload definitions}
\label{app:w-sensitivity}

The main paper uses $W_{\mathrm{poi}}$, but the ranking conclusions are stable under reasonable alternatives.  \Cref{tab:w-sensitivity} reports sensitivity results.  $W_{\mathrm{needle}}=\sum_{t\in\mathcal{N}}(k_t+m_t)$ assumes oracle knowledge of needle identity without condition checking.  Adding $N$ or $2N$ introduces explicit per-statement parsing costs.  All alternatives preserve the main result: overhead rankings are highly stable and the approximately nine-fold spread in $\bar{VO}$ remains.

\begin{table}[H]
\centering
\small
\begin{tabular}{lccc}
\toprule
Workload definition & Spearman $\rho$ vs. $W_{\mathrm{poi}}$ & Max rank shift & $\bar{VO}$ range \\
\midrule
$W_{\mathrm{poi}}$ & -- & -- & $9.29\times$ \\
$W_{\mathrm{needle}}$ & 0.998 & 1 & $9.23\times$ \\
$W_{\mathrm{poi}}+N$ & 0.994 & 2 & $9.13\times$ \\
$W_{\mathrm{poi}}+2N$ & 0.988 & 2 & $9.04\times$ \\
\bottomrule
\end{tabular}
\caption{Sensitivity of native CogniLoad overhead rankings to alternative workload definitions.}
\label{tab:w-sensitivity}
\end{table}

\subsection{Native workload sensitivity on GSM8K, ProofWriter, and ZebraLogic}
\label{app:transfer-workload-sensitivity}

\paragraph{GSM8K.}
Gold-step count is interpretable but coarse, so we also compute an arithmetic-operation workload from the annotated gold solution.  The operation workload has 17 unique values rather than 9 and improves agreement with solver-derived workload, especially for coupling.

\begin{table}[H]
\centering
\scriptsize
\setlength{\tabcolsep}{3pt}
\begin{tabular}{@{}lrrrrr@{}}
\toprule
Workload
& \makecell{Unique\\values}
& \makecell{Retained\\instances}
& \(\rho_{\vo}\)
& \(\rho_{\kappa}\)
& \makecell{Max \(\vo\)\\rank shift\\vs. step} \\
\midrule
\(W_{\mathrm{step}}\) & 9  & 1319 & 0.938 & 0.358 & 0 \\
\(W_{\mathrm{op}}\)   & 17 & 1319 & 0.965 & 0.644 & 2 \\
\(W_{\mathrm{mix}}\)  & 22 & 1319 & 0.938 & 0.429 & 0 \\
\bottomrule
\end{tabular}
\caption{GSM8K native-workload sensitivity. Here \(\rho_{\vo}=\rho(\vo,\vo_{\mathrm{solv}})\) and \(\rho_{\kappa}=\rho(\kappa,\kappa_{\mathrm{solv}})\). Arithmetic-operation count provides a finer workload scale and improves agreement with the solver-derived scale.}
\label{tab:gsm8k-workload-sensitivity}
\end{table}


\paragraph{ProofWriter.}
The perfect $\vo$ agreement under proof-depth workload is not a tied-rank artifact.  Although proof depth has only six values and each depth has 500 instances, both Spearman and Kendall $\tau_b$ equal 1.000 for $\vo$, with maximum rank difference zero.  Coupling is similar but not identical: Spearman is 0.925 and Kendall $\tau_b$ is 0.758.

\begin{table}[H]
\centering
\small
\setlength{\tabcolsep}{6pt}
\begin{tabular}{@{}lr@{}}
\toprule
Metric & Value \\
\midrule
Unique depths & 6 \\
Largest tie block & 500 \\
Coverage & 1.000 \\
Spearman \(\vo\) & 1.000 \\
Kendall \(\vo\) & 1.000 \\
Max \(\vo\) rank difference & 0 \\
Spearman \(\kappa\) & 0.925 \\
Kendall \(\kappa\) & 0.758 \\
\bottomrule
\end{tabular}
\caption{ProofWriter tie audit for native-versus-solver workload agreement.}
\label{tab:proofwriter-tie-audit}
\end{table}


\paragraph{ZebraLogic.}
ZebraLogic overhead rankings are invariant across the tested deterministic structural workload variants.  This indicates that the main ZebraLogic mean-overhead conclusions do not depend on a single arbitrary clue-counting rule.

\begin{table}[H]
\centering
\small
\begin{tabular}{lcccc}
\toprule
Comparison & $\rho(\vo)$ & $\rho(\kappa)$ & Max $\vo$ rank shift & Max $\kappa$ rank shift \\
\midrule
$W_{\mathrm{atom}}$ vs. $W_{\mathrm{weighted}}$ & 1.000 & 0.996 & 0 & 1 \\
$W_{\mathrm{atom}}$ vs. $W_{\mathrm{grid}}$     & 1.000 & 0.978 & 0 & 2 \\
\bottomrule
\end{tabular}
\caption{ZebraLogic structural-workload sensitivity.}
\label{tab:zebralogic-workload-sensitivity}
\end{table}

\subsection{Extending workload normalization beyond these benchmarks}
\label{app:generalization}

The framework can be applied wherever an evaluator can define an instance-level workload scale.  Examples include constraint counts for ZebraLogic-style puzzles, supporting-fact counts for retrieval-and-reasoning tasks, relational path length for CLUTRR-style tasks, proof depth for theorem/proof benchmarks, and reference-solution step count for mathematical word problems.  These scales differ in precision: generator-recorded operation counts are usually cleaner than reference-solution length.  The native/solver-derived distinction is meant to make that uncertainty explicit rather than hide it.

For unstructured benchmarks, approximate native scales may include reference solution length, number of entities, number of intermediate quantities, or number of required evidence passages.  Such proxies should be validated in the same way we validate the solver-derived scale here: by checking whether conclusions are stable under alternative plausible workload definitions.

\section{Additional cross-benchmark validation}
\label{app:cross-benchmark}

\Cref{tab:cross-cogniload-pearson,tab:cross-cogniload-spearman} report CogniLoad-to-transfer correlations for raw metric values and rank orderings.  Workload-layer quantities use solver-derived workload so that the same workload construction is available for all four benchmarks.  The trace column uses the 10 shared trace-visible models.

\begin{table}[H]
\caption{Pearson correlations of raw metric values against CogniLoad.  Outcome and workload columns use the 14 shared open-weight models; $q_{\mathrm{trace}}$ uses the 10 shared trace-visible models.}
\label{tab:cross-cogniload-pearson}
\centering
\small
\setlength{\tabcolsep}{4pt}
\begin{tabular}{lrrrrrrrr}
\toprule
Benchmark & $E_0$ & Acc & $r_{\mathrm{ctx}}$ & $r_{\mathrm{logic}}$ & $\mathbb{E}[T]$ & $\bar{VO}$ & $\kappa$ & $q_{\mathrm{trace}}$ \\
\midrule
GSM8K       & 0.776 & 0.690 & $-$0.024 & 0.704 & 0.558 & 0.328 & $-$0.583 & $-$0.252 \\
ProofWriter & 0.599 & 0.678 & $-$0.067 & 0.679 & 0.684 & 0.405 & $-$0.246 & 0.176 \\
ZebraLogic  & 0.428 & 0.362 & 0.225  & 0.294 & 0.736 & 0.897 & 0.577  & 0.217 \\
\bottomrule
\end{tabular}
\end{table}

\begin{table}[H]
\caption{Spearman rank correlations against CogniLoad.  Outcome and workload columns use the 14 shared open-weight models; $q_{\mathrm{trace}}$ uses the 10 shared trace-visible models.}
\label{tab:cross-cogniload-spearman}
\centering
\small
\setlength{\tabcolsep}{4pt}
\begin{tabular}{lrrrrrrrr}
\toprule
Benchmark & $E_0$ & Acc & $r_{\mathrm{ctx}}$ & $r_{\mathrm{logic}}$ & $\mathbb{E}[T]$ & $\bar{VO}$ & $\kappa$ & $q_{\mathrm{trace}}$ \\
\midrule
GSM8K       & 0.859 & 0.530 & $-$0.111 & 0.723 & 0.437 & 0.547 & $-$0.591 & $-$0.152 \\
ProofWriter & 0.679 & 0.367 & 0.243  & 0.503 & 0.622 & 0.613 & $-$0.165 & 0.200 \\
ZebraLogic  & 0.534 & 0.292 & 0.288  & 0.512 & 0.631 & 0.811 & 0.389  & 0.321 \\
\bottomrule
\end{tabular}
\end{table}

\Cref{tab:native-pairwise,tab:solver-pairwise} provide the full pairwise workload comparison under native and solver-derived workload scales.  The solver-derived scale yields more consistently positive $\bar{VO}$ correlations, which is why the main paper uses it for the benchmark-uniform workload analysis.  The solver-derived table also reports accuracy correlations side-by-side with $E_0$ and workload-normalized overhead.

\begin{table}[H]
\caption{Pairwise Spearman correlations under native workload.}
\label{tab:native-pairwise}
\centering
\small
\begin{tabular}{lccc}
\toprule
Benchmark pair & $\rho(E_0)$ & $\rho(\bar{VO}_{\mathrm{nat}})$ & $\rho(\kappa_{\mathrm{nat}})$ \\
\midrule
CogniLoad--GSM8K       & 0.859 & 0.547 & $-$0.591 \\
CogniLoad--ProofWriter & 0.679 & 0.613 & $-$0.165 \\
CogniLoad--ZebraLogic  & 0.534 & 0.807 & 0.389 \\
GSM8K--ProofWriter     & 0.802 & 0.670 & 0.284 \\
GSM8K--ZebraLogic      & 0.521 & 0.363 & $-$0.073 \\
ProofWriter--ZebraLogic& 0.547 & 0.358 & 0.354 \\
\bottomrule
\end{tabular}
\end{table}

\begin{table}[H]
\caption{Pairwise Spearman correlations under solver-derived workload.}
\label{tab:solver-pairwise}
\centering
\small
\setlength{\tabcolsep}{3.5pt}
\begin{tabular}{lcccc}
\toprule
Benchmark pair & $\rho(E_0)$ & $\rho(\mathrm{Acc})$ & $\rho(\bar{VO}_{\mathrm{solv}})$ & $\rho(\kappa_{\mathrm{solv}})$ \\
\midrule
CogniLoad--GSM8K       & 0.881 & 0.587 & 0.705 & $-$0.354 \\
CogniLoad--ProofWriter & 0.705 & 0.332 & 0.758 & $-$0.275 \\
CogniLoad--ZebraLogic  & 0.499 & 0.354 & 0.771 & $-$0.231 \\
GSM8K--ProofWriter     & 0.780 & 0.304 & 0.785 & 0.807 \\
GSM8K--ZebraLogic      & 0.424 & 0.777 & 0.604 & 0.411 \\
ProofWriter--ZebraLogic& 0.582 & 0.288 & 0.429 & 0.516 \\
\bottomrule
\end{tabular}
\end{table}

\subsection{Cross-benchmark rank-correlation uncertainty}
\label{app:cross-benchmark-uncertainty}

For the solver-derived pairwise table, \Cref{tab:transfer-bootstrap} reports nonparametric bootstrap intervals over the 14 shared model ranks.  Accuracy has the lowest median full-sample correlation ($0.343$) and the widest median interval among the main transfer quantities.  $E_0$ and $\bar{VO}_{\mathrm{solv}}$ exceed accuracy in five of six benchmark pairs; GSM8K--ZebraLogic is the exception because its accuracy ranking is itself highly stable across the two benchmarks.

\begin{table}[H]
\caption{Bootstrap intervals for pairwise Spearman rank correlations in the benchmark-uniform solver-derived analysis.  Entries are full-sample $\rho$ with 2.5--97.5\% bootstrap intervals in brackets.}
\label{tab:transfer-bootstrap}
\centering
\scriptsize
\setlength{\tabcolsep}{2pt}
\begin{tabular}{lcccc}
\toprule
Benchmark pair & $\rho(E_0)$ & $\rho(\mathrm{Acc})$ & $\rho(\bar{VO}_{\mathrm{solv}})$ & $\rho(\kappa_{\mathrm{solv}})$ \\
\midrule
CogniLoad--GSM8K       & $0.881\,[0.724,0.984]$ & $0.587\,[0.042,0.931]$ & $0.705\,[0.351,0.951]$ & $-0.354\,[-0.716,0.155]$ \\
CogniLoad--ProofWriter & $0.705\,[0.414,0.908]$ & $0.332\,[-0.269,0.702]$ & $0.758\,[0.484,0.916]$ & $-0.275\,[-0.670,0.255]$ \\
CogniLoad--ZebraLogic  & $0.499\,[-0.095,0.936]$ & $0.354\,[-0.248,0.881]$ & $0.771\,[0.494,0.943]$ & $-0.231\,[-0.848,0.291]$ \\
GSM8K--ProofWriter     & $0.780\,[0.459,0.955]$ & $0.304\,[-0.359,0.712]$ & $0.785\,[0.457,0.944]$ & $0.807\,[0.539,0.943]$ \\
GSM8K--ZebraLogic      & $0.424\,[-0.085,0.828]$ & $0.777\,[0.571,0.915]$ & $0.604\,[0.238,0.902]$ & $0.411\,[-0.132,0.816]$ \\
ProofWriter--ZebraLogic& $0.582\,[0.140,0.875]$ & $0.288\,[-0.295,0.758]$ & $0.429\,[0.018,0.779]$ & $0.516\,[-0.002,0.876]$ \\
\bottomrule
\end{tabular}
\end{table}

\begin{table}[H]
\caption{Calibration of native and solver-derived workload rankings.  High $\bar{VO}$ correlations support the solver-derived fallback for mean overhead.}
\label{tab:workload-calibration-app}
\centering
\small
\begin{tabular}{lccc}
\toprule
Benchmark / native scale & Unique $W$ values & $\rho(\bar{VO}_{\mathrm{nat}},\bar{VO}_{\mathrm{solv}})$ & $\rho(\kappa_{\mathrm{nat}},\kappa_{\mathrm{solv}})$ \\
\midrule
CogniLoad $W_{\mathrm{poi}}$ & -- & 0.982 & 0.933 \\
GSM8K $W_{\mathrm{step}}$ & 9 & 0.938 & 0.358 \\
GSM8K $W_{\mathrm{op}}$ & 17 & 0.965 & 0.644 \\
ProofWriter $W_{\mathrm{depth}}$ & 6 & 1.000 & 0.925 \\
ZebraLogic $W_{\mathrm{atom}}$ & 29 & 0.969 & 0.934 \\
\bottomrule
\end{tabular}
\end{table}

\subsection{What should and should not transfer}
\label{app:what-transfer}
Our protocol separates quantities with different intended scopes.  $E_0$ and $\vo$ are natural cross-model comparative statistics.  They ask how many correct answers are bought by a token budget and how many tokens are spent per declared unit of work.  $\kappa$ is different: it describes how overhead varies across the workload geometry of one benchmark.  A model can have stable mean overhead across benchmarks while having different coupling because GSM8K steps, ProofWriter depths, CogniLoad condition checks, and ZebraLogic constraints are not commensurable units.  Treating $\kappa$ as local preserves its diagnostic value without overstating its transferability.

\section{Full model list and CogniLoad results}
\label{app:models-results}

\subsection{Model list and evaluation coverage}

\cref{tab:model-list-full} reports the full model list with realized evaluation coverage. 

\begin{table}[H]
\centering
\small
\setlength{\tabcolsep}{4pt}
\begin{tabular}{llrl}
\toprule
Model & Category & Parameters & Experiments analyzed \\
\midrule
o3-2025-04-16 & OpenAI API & Unknown & 1,398 \\
gpt-5-2025-08-07 & OpenAI API & Unknown & 1,400 \\
gpt-5-mini & OpenAI API & Unknown & 1,399 \\
gpt-5-nano & OpenAI API & Unknown & 1,399 \\
o4-mini & OpenAI API & Unknown & 1,398 \\
gemini-2.5-pro & Gemini API & Unknown & 1,400 \\
gemini-2.5-flash & Gemini API & Unknown & 1,400 \\
gemini-2.5-flash-lite & Gemini API & Unknown & 1,400 \\
glm-4.5-air & Open-weight/API-served & 355B MoE & 1,397 \\
DeepSeek-R1-0528 & Open-weight/API-served & 685B MoE & 1,339 \\
DeepSeek-R1-Distill-Llama-70B & Open-weight/local & 70B & 14,000 \\
DeepSeek-R1-Distill-Qwen-32B & Open-weight/local & 32B & 14,000 \\
DeepSeek-R1-Distill-Qwen-7B & Open-weight/local & 7B & 14,000 \\
DeepSeek-R1-Distill-Qwen-1.5B & Open-weight/local & 1.5B & 14,000 \\
QwQ-32B & Open-weight/local & 32B & 14,000 \\
GLM-Z1-32B & Open-weight/local & 32B & 14,000 \\
EXAONE-Deep-32B & Open-weight/local & 32B & 14,000 \\
Qwen3-32B & Open-weight/local & 32B & 14,000 \\
Qwen3-30B-A3B & Open-weight/local & 30B MoE & 14,000 \\
Qwen3-8B & Open-weight/local & 8B & 13,999 \\
Qwen3-1.7B & Open-weight/local & 1.7B & 13,998 \\
Phi-4-reasoning-plus & Open-weight/local & 14B & 14,000 \\
Phi-4-reasoning & Open-weight/local & 14B & 14,000 \\
Phi-4-mini-reasoning & Open-weight/local & 3.8B & 14,000 \\
MiMo-7B-RL & Open-weight/local & 7B & 13,929 \\
\bottomrule
\end{tabular}
\caption{Full CogniLoad model list with realized evaluation coverage.}
\label{tab:model-list-full}
\end{table}

\subsection{Outcome-level leaderboard}

\cref{tab:leaderboard-full} reports the outcome-level CogniLoad leaderboard.

\begin{table}[H]
\caption{Outcome-level CogniLoad leaderboard.  For compactness, the table reports $100E_0$; e.g., 15.25 corresponds to $E_0=0.1525$ correct answers per 1,000 generated output tokens.  Acc, $r_{\mathrm{ctx}}$, and $r_{\mathrm{logic}}$ are percentages.}
\label{tab:leaderboard-full}
\centering
\small
\setlength{\tabcolsep}{3.5pt}
\begin{tabular}{lrrrrr}
\toprule
Model & $100E_0$ & Acc & $\mathbb{E}[T]$ & $r_{\mathrm{ctx}}$ & $r_{\mathrm{logic}}$ \\
\midrule
o3 & 15.25 & 89 & 5869 & 98 & 91 \\
GPT-5 & 13.41 & 92 & 6897 & 98 & 95 \\
DS-R1D-Llama-70B & 12.57 & 58 & 4592 & 100 & 58 \\
o4-mini & 12.27 & 76 & 6232 & 98 & 78 \\
GLM-Z1-32B & 11.94 & 48 & 3980 & 100 & 48 \\
Qwen3-32B & 11.47 & 50 & 4383 & 100 & 50 \\
Gemini-2.5-flash-lite & 11.36 & 38 & 3339 & 99 & 38 \\
GPT-5-mini & 8.91 & 79 & 8878 & 97 & 82 \\
Qwen3-30B-A3B & 8.77 & 45 & 5105 & 99 & 45 \\
DS-R1D-Qwen-32B & 8.03 & 51 & 6393 & 100 & 51 \\
DS-R1-0528 & 7.18 & 68 & 9476 & 100 & 68 \\
Qwen3-8B & 6.45 & 41 & 6405 & 96 & 43 \\
QwQ-32B & 6.18 & 50 & 8168 & 97 & 52 \\
GPT-5-nano & 5.93 & 46 & 7792 & 100 & 46 \\
GLM-4.5-air & 5.91 & 59 & 9935 & 94 & 63 \\
Qwen3-1.7B & 5.30 & 21 & 3949 & 100 & 21 \\
Gemini-2.5-pro & 5.21 & 75 & 14416 & 77 & 97 \\
Gemini-2.5-flash & 4.62 & 66 & 14332 & 72 & 92 \\
Phi-4-reasoning & 4.24 & 51 & 12013 & 88 & 58 \\
DS-R1D-Qwen-7B & 3.84 & 26 & 6858 & 100 & 26 \\
Phi-4-mini-reasoning & 3.21 & 26 & 8005 & 100 & 26 \\
EXAONE-Deep-32B & 3.14 & 35 & 11060 & 100 & 35 \\
Phi-4-reasoning-plus & 3.08 & 46 & 15046 & 79 & 58 \\
MiMo-7B-RL & 2.99 & 6 & 2048 & 100 & 6 \\
DS-R1D-Qwen-1.5B & 1.26 & 19 & 14690 & 94 & 20 \\
\bottomrule
\end{tabular}
\end{table}

\subsection{Native workload-normalized verbosity on CogniLoad}

\cref{tab:vo-full} reports the native workload-normalized verbosity on CogniLoad using $W_{\mathrm{poi}}$.

\begin{table}[H]
\caption{Native workload-normalized verbosity on CogniLoad using $W_{\mathrm{poi}}$.}
\label{tab:vo-full}
\centering
\small
\setlength{\tabcolsep}{4pt}
\begin{tabular}{lrrrr}
\toprule
Model & Acc & $\mathbb{E}[T]$ & $\bar{VO}_{\mathrm{nat}}$ & $\kappa_{\mathrm{nat}}$ \\
\midrule
MiMo-7B-RL & 0.06 & 2048 & 13.67 & 0.30 \\
Gemini-2.5-flash-lite & 0.38 & 3339 & 14.67 & 0.46 \\
o3 & 0.89 & 5869 & 19.84 & 0.60 \\
DS-R1D-Llama-70B & 0.58 & 4592 & 20.03 & 0.47 \\
GLM-Z1-32B & 0.48 & 3980 & 21.27 & 0.38 \\
Qwen3-32B & 0.50 & 4383 & 22.79 & 0.39 \\
GPT-5 & 0.93 & 6897 & 24.17 & 0.58 \\
o4-mini & 0.76 & 6232 & 24.37 & 0.52 \\
DS-R1D-Qwen-32B & 0.51 & 6393 & 25.23 & 0.52 \\
Qwen3-30B-A3B & 0.45 & 5105 & 28.31 & 0.37 \\
Qwen3-1.7B & 0.21 & 3949 & 29.44 & 0.27 \\
GPT-5-mini & 0.79 & 8884 & 31.84 & 0.57 \\
DS-R1D-Qwen-7B & 0.26 & 6858 & 35.35 & 0.40 \\
Qwen3-8B & 0.41 & 6405 & 35.44 & 0.37 \\
GPT-5-nano & 0.46 & 7798 & 38.76 & 0.41 \\
QwQ-32B & 0.51 & 8168 & 39.64 & 0.42 \\
DS-R1-0528 & 0.68 & 9300 & 45.17 & 0.42 \\
GLM-4.5-air & 0.59 & 9935 & 46.30 & 0.44 \\
Gemini-2.5-flash & 0.66 & 14332 & 49.08 & 0.59 \\
Phi-4-reasoning & 0.51 & 12013 & 50.46 & 0.48 \\
Gemini-2.5-pro & 0.75 & 14416 & 53.01 & 0.55 \\
Phi-4-mini-reasoning & 0.26 & 8005 & 53.10 & 0.31 \\
Phi-4-reasoning-plus & 0.46 & 15046 & 66.45 & 0.46 \\
EXAONE-Deep-32B & 0.35 & 11060 & 69.30 & 0.33 \\
DS-R1D-Qwen-1.5B & 0.19 & 14690 & 127.03 & 0.24 \\
\bottomrule
\end{tabular}
\end{table}

\begin{figure}[H]
\centering
\includegraphics[width=0.62\linewidth]{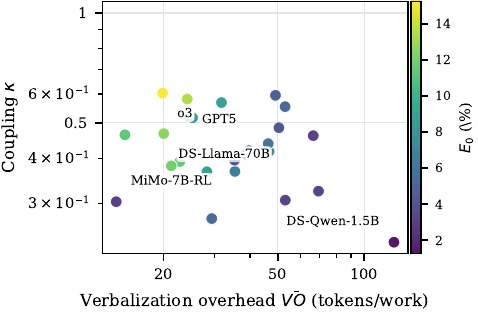}
\caption{Mean verbalization overhead versus workload coupling on CogniLoad.  Most cross-model verbosity differences are explained by $\bar{VO}$ rather than $\kappa$.}
\label{fig:vo-kappa-app}
\end{figure}

\cref{fig:vo-kappa-app} plots the mean verbalization overhead versus workload coupling on CogniLoad.

\section{Compression trace-quality implementation}
\label{app:trace-impl}

For trace-visible runs, we extract the visible reasoning text using benchmark-specific parsers: chain-of-thought text before the final answer for GSM8K and ProofWriter, the JSON \texttt{reasoning} field or \texttt{<think>} block for ZebraLogic, and the recorded visible reasoning trace for CogniLoad.  Hidden-reasoning API systems are excluded from the trace layer because their internal trace is not observable.

The trace-quality signal is the bounded zlib compression ratio
\begin{equation}
\sigma_c(I)=\min\left(1,\frac{|\mathrm{zlib}(\mathrm{trace}(I))|}{|\mathrm{trace}(I)|}\right).
\end{equation}
Lower values indicate more redundant visible text; higher values indicate less compressible, higher-density text.  We then set $T_{\mathrm{sig}}(I)=T(I)\sigma_c(I)$ and use the same workload factorization as the main text.  This diagnostic is deterministic, annotation-free, and does not require a benchmark ontology, which is why it is available for all four benchmarks whenever traces are visible.  It is not a semantic proof checker: a high-compression signal can still be wrong, and a low-compression signal can still contain some useful reasoning.

\begin{table}[H]
\centering
\small
\begin{tabular}{lrrrr}
\toprule
Model & $q_{\mathrm{trace}}$ & $\bar{VO}_{\mathrm{sig}}$ & $\kappa_{\mathrm{sig}}$ & Mean $\sigma_c$ \\
\midrule
DS-R1D-Llama-70B & 0.182 & 4.03 & 0.42 & 0.217 \\
GLM-Z1-32B & 0.234 & 5.22 & 0.36 & 0.250 \\
DS-R1D-Qwen-32B & 0.166 & 4.86 & 0.44 & 0.199 \\
DS-R1-0528 & 0.208 & 9.67 & 0.41 & 0.215 \\
QwQ-32B & 0.194 & 8.79 & 0.37 & 0.213 \\
GLM-4.5-air & 0.255 & 10.72 & 0.48 & 0.279 \\
Phi-4-reasoning & 0.144 & 8.27 & 0.43 & 0.158 \\
DS-R1D-Qwen-7B & 0.108 & 5.27 & 0.29 & 0.199 \\
Phi-4-mini-reasoning & 0.231 & 12.16 & 0.31 & 0.237 \\
EXAONE-Deep-32B & 0.214 & 15.35 & 0.31 & 0.221 \\
Phi-4-reasoning-plus & 0.149 & 11.03 & 0.41 & 0.157 \\
DS-R1D-Qwen-1.5B & 0.052 & 7.00 & 0.22 & 0.175 \\
\bottomrule
\end{tabular}
\caption{Compression trace-quality statistics for CogniLoad trace-visible models.  $q_{\mathrm{trace}}$ is token-weighted, while mean $\sigma_c$ averages the per-instance compression ratio.}
\label{tab:layer3-stats}
\end{table}

\begin{figure}[H]
\centering
\includegraphics[width=0.98\linewidth]{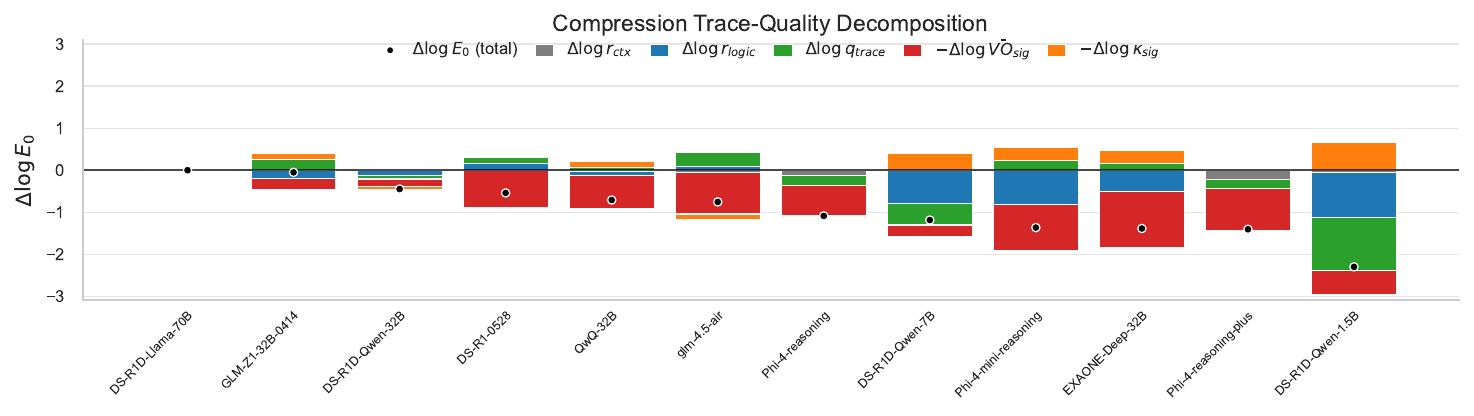}
\caption{Compression-trace decomposition for the 12 trace-visible CogniLoad models, using DeepSeek-R1-Distill-Llama-70B as the reference.  The trace layer separates redundancy-dominated collapse from high compressed-signal overhead.}
\label{fig:trace-decomp-app}
\end{figure}

\section{Compression trace quality across benchmarks}
\label{app:compression}

Because compression needs only visible trace text, the same statistic can be computed on GSM8K, ProofWriter, and ZebraLogic.  Its cross-benchmark rank agreement is weaker than $E_0$ or $\bar{VO}$: against CogniLoad, Spearman correlations are $-0.152$ on GSM8K, $0.200$ on ProofWriter, and $0.321$ on ZebraLogic over the 10 shared trace-visible models.  We therefore interpret $q_{\mathrm{trace}}$ as a local diagnostic for redundancy and information density, not as a benchmark-stable model-quality ranking.

\begin{table}[H]
\centering
\small
\setlength{\tabcolsep}{4pt}
\begin{tabular}{lrrrr}
\toprule
Benchmark & Shared models & $\rho(E_0)$ & $\rho(\bar{VO}_{\mathrm{sig}})$ & $\rho(q_{\mathrm{trace}})$ \\
\midrule
GSM8K & 10 & 0.806 & 0.879 & $-$0.152 \\
ProofWriter & 10 & 0.685 & 0.636 & 0.200 \\
ZebraLogic & 10 & 0.358 & 0.818 & 0.321 \\
\bottomrule
\end{tabular}
\caption{CogniLoad-to-transfer Spearman rank correlations for the compression trace layer.}
\label{tab:compression-quality}
\end{table}

\section{Robustness checks}
\label{app:robustness}

\subsection{Tokens as a compute proxy}

Generated tokens are observable and deployment-relevant, but they are not equivalent to FLOPs, latency, energy, or dollar cost.  Tokenizers also differ across providers.  As a robustness check, we recompute efficiency using generated characters instead of generated tokens.  Character-normalized and token-normalized $E_0$ rankings have Spearman correlation $0.98$, with maximum rank displacement of one position.  Tokenization idiosyncrasies therefore do not drive the reported ranking conclusions.

\subsection{Alternative cost columns}
\label{app:alternative-costs}

Visible generated tokens are the primary metric because they are observable across open and closed systems.  We nevertheless audit alternative cost columns where available.  On GSM8K, ProofWriter, and ZebraLogic, replacing output tokens with total tokens leaves efficiency rankings nearly unchanged: Spearman correlation with $E_0$ ranges from $0.987$ to $1.000$.  Replacing tokens with elapsed time gives a different view: correlations range from $0.270$ on GSM8K to $0.899$ on ProofWriter.  On CogniLoad local runs, estimated GPU-hour and MI250X-card-hour efficiency correlate $0.325$ with visible-token efficiency.  These results support reporting visible-token efficiency as a common output-channel metric, while treating latency, hardware time, energy, and dollar cost as parallel deployment-cost audits.

\begin{table}[H]
\centering
\small
\setlength{\tabcolsep}{4pt}
\begin{tabular}{llcc}
\toprule
Group & Alternative cost & Models & $\rho$ with visible-token $E_0$ \\
\midrule
GSM8K local & total tokens & 14 & 0.987 \\
GSM8K local & elapsed seconds & 14 & 0.270 \\
ProofWriter local & total tokens & 14 & 1.000 \\
ProofWriter local & elapsed seconds & 14 & 0.899 \\
ZebraLogic local & total tokens & 14 & 0.996 \\
ZebraLogic local & elapsed seconds & 14 & 0.820 \\
CogniLoad local & estimated GPU hours & 15 & 0.325 \\
CogniLoad local & equivalent MI250X card-hours & 15 & 0.325 \\
\bottomrule
\end{tabular}
\caption{Alternative cost columns.  Total-token rankings remain close to visible-output-token rankings, while elapsed-time and hardware-resource views can differ.}
\label{tab:alternative-costs}
\end{table}

\subsection{Nested-subset uncertainty for CogniLoad}

Cost-constrained API models use a fixed 10-instance subset per CogniLoad configuration, whereas local models use 100 instances per configuration.  The smaller suite is nested in the larger suite, so we can simulate the API sampling regime by repeatedly subsampling local-model runs.  Across 1000 nested-subset resamples, the 10-instance-per-configuration subset recovers full-run values and rankings well.  Median relative error is $2.0\%$ for $E_0$, $1.2\%$ for $\vo$, $1.6\%$ for accuracy, $1.0\%$ for $\kappa$, $0.1\%$ for $r_{\mathrm{ctx}}$, and $1.6\%$ for $r_{\mathrm{logic}}$.  The corresponding rank-stability medians are $0.995$ for $E_0$, $1.000$ for $\vo$, $0.982$ for accuracy, $0.989$ for $\kappa$, $0.981$ for $r_{\mathrm{ctx}}$, and $0.984$ for $r_{\mathrm{logic}}$.

\begin{table}[H]
\centering
\small
\begin{tabular}{lccc}
\toprule
Metric & Median relative error & 95th percentile error & Median rank $\rho$ \\
\midrule
$E_0$ & 0.020 & 0.070 & 0.995 \\
$\vo$ & 0.012 & 0.065 & 1.000 \\
Accuracy & 0.016 & 0.059 & 0.982 \\
$\kappa$ & 0.010 & 0.052 & 0.989 \\
$r_{\mathrm{ctx}}$ & 0.001 & 0.010 & 0.981 \\
$r_{\mathrm{logic}}$ & 0.016 & 0.058 & 0.984 \\
\bottomrule
\end{tabular}
\caption{Nested-subset uncertainty for CogniLoad under the 10-instance-per-configuration sampling regime.}
\label{tab:nested-subset-uncertainty}
\end{table}
\subsection{Selection effects in $\kappa$}
\label{app:selection}

A potential concern is that $\kappa<1$ could arise because models abandon hard problems early, rather than because they compress reasoning on higher-workload instances.  We checked this by comparing token counts for correct and incorrect completions, excluding truncations.  Across the 25 CogniLoad models, incorrect answers produce more tokens than correct answers on average (mean ratio $1.23\pm0.25$).  The strongest models show the largest over-investment in failures: GPT-5 produces 86\% more tokens on incorrect completions than correct ones, o3 produces 67\% more, and GPT-5-mini produces 58\% more.  Only two models show ratios below one.

\begin{table}[H]
\centering
\small
\setlength{\tabcolsep}{4pt}
\begin{tabular}{lrrrl}
\toprule
Model & $\mathbb{E}[T\mid\mathrm{correct}]$ & $\mathbb{E}[T\mid\mathrm{incorrect}]$ & Ratio & Pattern \\
\midrule
GPT-5 & 6,370 & 11,856 & 1.86 & Over-invest \\
o3 & 5,351 & 8,936 & 1.67 & Over-invest \\
GPT-5-mini & 7,751 & 12,280 & 1.58 & Over-invest \\
Gemini-2.5-flash & 10,376 & 15,437 & 1.49 & Over-invest \\
o4-mini & 5,532 & 7,814 & 1.41 & Over-invest \\
Phi-4-reasoning & 9,130 & 12,856 & 1.41 & Over-invest \\
Phi-4-reasoning-plus & 11,217 & 15,603 & 1.39 & Over-invest \\
DS-R1D-Qwen-32B & 5,385 & 7,421 & 1.38 & Over-invest \\
Gemini-2.5-pro & 11,817 & 15,304 & 1.30 & Over-invest \\
DS-R1D-Llama-70B & 4,155 & 5,188 & 1.25 & Over-invest \\
DS-R1-0528 & 8,943 & 10,698 & 1.20 & Over-invest \\
GLM-4.5-air & 8,480 & 10,212 & 1.20 & Over-invest \\
QwQ-32B & 7,145 & 8,474 & 1.19 & Over-invest \\
Qwen3-32B & 3,994 & 4,772 & 1.19 & Over-invest \\
GLM-Z1-32B & 3,687 & 4,229 & 1.15 & Moderate \\
Qwen3-8B & 5,262 & 5,932 & 1.13 & Moderate \\
DS-R1D-Qwen-7B & 6,245 & 7,013 & 1.12 & Moderate \\
GPT-5-nano & 7,323 & 8,195 & 1.12 & Moderate \\
Qwen3-30B-A3B & 4,665 & 5,013 & 1.07 & Moderate \\
DS-R1D-Qwen-1.5B & 10,362 & 10,862 & 1.05 & Balanced \\
EXAONE-Deep-32B & 10,985 & 11,026 & 1.00 & Balanced \\
MiMo-7B-RL & 2,045 & 2,048 & 1.00 & Balanced \\
Phi-4-mini-reasoning & 7,969 & 7,969 & 1.00 & Balanced \\
Qwen3-1.7B & 4,303 & 3,853 & 0.90 & Shorter failures \\
Gemini-2.5-flash-lite & 3,581 & 2,798 & 0.78 & Shorter failures \\
\bottomrule
\end{tabular}
\caption{Token counts by correctness, excluding budget truncations.  The sublinear $\kappa$ pattern is not explained solely by early abandonment on hard failures.}
\label{tab:tokens-by-outcome-full}
\end{table}

\subsection{Bootstrap confidence intervals}
\label{app:decomp-ci}

We compute nonparametric bootstrap intervals for the decomposition terms using 500 resamples over instances.  Typical 95\% confidence interval widths are 0.03--0.05 in log space; differences larger than 0.1 in point estimates are therefore generally well separated.  Examples include the logic robustness gap between o3 and DeepSeek-R1-0528 ($\Delta\log r_{\mathrm{logic}}=-0.29$), the EXAONE-Deep-32B overhead contribution ($-1.25$), and the Gemini~2.5~Flash context deficit ($-0.31$).

\begin{table}[H]
\centering
\small
\begin{tabular}{lccc}
\toprule
Model & $\Delta\log r_{\mathrm{logic}}$ & $\Delta\log r_{\mathrm{ctx}}$ & $-\Delta\log\bar{VO}$ \\
\midrule
o3 (ref) & 0.00 & 0.00 & 0.00 \\
GPT-5 & +0.04 & $-$0.01 & $-$0.20 \\
o4-mini & $-$0.16 & 0.00 & $-$0.21 \\
DeepSeek-R1-0528 & $-$0.29 & +0.02 & $-$0.82 \\
DS-R1D-Llama-70B & $-$0.46 & +0.02 & $-$0.01 \\
QwQ-32B & $-$0.56 & $-$0.02 & $-$0.69 \\
Qwen3-32B & $-$0.59 & +0.02 & $-$0.14 \\
GLM-Z1-32B & $-$0.65 & +0.02 & $-$0.07 \\
GPT-5-nano & $-$0.68 & +0.02 & $-$0.67 \\
EXAONE-Deep-32B & $-$0.96 & +0.01 & $-$1.25 \\
DS-R1D-Qwen-7B & $-$1.24 & +0.02 & $-$0.58 \\
Phi-4-mini-reasoning & $-$1.26 & +0.01 & $-$0.98 \\
Qwen3-1.7B & $-$1.47 & +0.02 & $-$0.39 \\
DS-R1D-Qwen-1.5B & $-$1.53 & $-$0.04 & $-$1.86 \\
MiMo-7B-RL & $-$2.70 & +0.01 & +0.37 \\
\midrule
\multicolumn{4}{l}{\textit{Context-limited examples:}} \\
Gemini-2.5-flash & +0.01 & $-$0.31 & $-$0.91 \\
Gemini-2.5-pro & +0.06 & $-$0.24 & $-$0.98 \\
Phi-4-reasoning-plus & $-$0.44 & $-$0.21 & $-$1.21 \\
\bottomrule
\end{tabular}
\caption{Selected decomposition terms relative to o3.  Bootstrap intervals are omitted for compactness but are typically narrow in log space.}
\label{tab:decomp-ci}
\end{table}

\subsection{Oracle early-stop upper bound}
\label{app:oracle-early-stop}

For each correct generation, we estimate the first prefix at which the answer extractor would already return the correct answer.  Replacing the full generation length with this prefix length gives an oracle upper bound on gains from perfect stopping, without changing any incorrect generations.  This is not an implementable stopping rule, because it uses the gold answer, but it measures how much observed inefficiency is attributable to text after the first correct answer appears.

\begin{table}[H]
\centering
\small
\begin{tabular}{lccc}
\toprule
Benchmark & Median gain & Max gain & Median wasted-tail fraction \\
\midrule
CogniLoad   & 1.298 & 2.694 & 0.922 \\
GSM8K       & 1.004 & 1.618 & 0.000 \\
ProofWriter & 1.004 & 1.309 & 0.000 \\
ZebraLogic  & 1.000 & 1.045 & 0.000 \\
\bottomrule
\end{tabular}
\caption{Oracle early-stop upper bound.  CogniLoad has substantial post-answer verbosity; the transfer benchmarks do not.}
\label{tab:oracle-early-stop}
\end{table}

\subsection{Truncation-loop audit}
\label{app:truncation-loop-audit}

We audit truncated generations by measuring repeated 4-grams and prompt-copy overlap in the final window of generated text.  No answer files were missing.  Truncation is concentrated in a small number of model/benchmark pairs.  At the benchmark level, median truncation rates are low, but the maximum model-level truncation rate reaches $85.9\%$ on ZebraLogic.

\begin{table}[H]
\centering
\small
\begin{tabular}{lcccc}
\toprule
Benchmark & Total truncated & Max truncation rate & Mean repeated tail & Mean prompt-copy tail \\
\midrule
CogniLoad   & 7568 & 0.278 & 0.504 & 0.780 \\
GSM8K       &  334 & 0.117 & 0.067 & 0.180 \\
ProofWriter &  573 & 0.094 & 0.082 & 0.138 \\
ZebraLogic  & 2166 & 0.859 & 0.287 & 0.561 \\
\bottomrule
\end{tabular}
\caption{Benchmark-level truncation-loop audit.  Repetition and prompt-copy fractions are computed on truncated tails.}
\label{tab:truncation-loop-audit}
\end{table}

\section{Resource accounting}
\label{app:resources}

\Cref{tab:resource-full} reports per-model resource figures.  Locally served open-weight models were run on LUMI using AMD MI250X cards with vLLM and 50 concurrent requests.  OpenAI and Gemini models were run through batch processing where available, reducing API costs by 50\%.  These numbers are included to make the scale of the evaluation auditable and to answer reviewer questions about computational cost.

\begin{table*}[h]
\centering
\small
\setlength{\tabcolsep}{3pt}
\begin{tabular}{llrrr}
\toprule
Model & Category & Experiments & MI250X card-hours & API USD \\
\midrule
o3-2025-04-16 & OpenAI & 1,398 & -- & 57 \\
gpt-5-2025-08-07 & OpenAI & 1,400 & -- & 56 \\
gpt-5-mini & OpenAI & 1,399 & -- & 21 \\
gpt-5-nano & OpenAI & 1,399 & -- & 3 \\
o4-mini & OpenAI & 1,398 & -- & 23 \\
gemini-2.5-pro & Gemini & 1,400 & -- & 307 \\
gemini-2.5-flash & Gemini & 1,400 & -- & 76 \\
gemini-2.5-flash-lite & Gemini & 1,400 & -- & 4 \\
glm-4.5-air & Open-weight/API & 1,397 & -- & 16 \\
DeepSeek-R1-0528 & Open-weight/API & 1,339 & -- & 30 \\
DS-R1D-Llama-70B & Open-weight/local & 14,000 & 260.78 & -- \\
DS-R1D-Qwen-32B & Open-weight/local & 14,000 & 320.46 & -- \\
DS-R1D-Qwen-7B & Open-weight/local & 14,000 & 473.70 & -- \\
DS-R1D-Qwen-1.5B & Open-weight/local & 14,000 & 1,728.90 & -- \\
QwQ-32B & Open-weight/local & 14,000 & 193.14 & -- \\
GLM-Z1-32B & Open-weight/local & 14,000 & 1,695.21 & -- \\
EXAONE-Deep-32B & Open-weight/local & 14,000 & 252.49 & -- \\
Qwen3-32B & Open-weight/local & 14,000 & 331.02 & -- \\
Qwen3-30B-A3B & Open-weight/local & 14,000 & 355.86 & -- \\
Qwen3-8B & Open-weight/local & 13,999 & 332.78 & -- \\
Qwen3-1.7B & Open-weight/local & 13,998 & 265.51 & -- \\
Phi-4-reasoning-plus & Open-weight/local & 14,000 & 300.53 & -- \\
Phi-4-reasoning & Open-weight/local & 14,000 & 186.05 & -- \\
Phi-4-mini-reasoning & Open-weight/local & 14,000 & 355.75 & -- \\
MiMo-7B-RL & Open-weight/local & 13,929 & 96.05 & -- \\
\midrule
Totals & & & 7,148.24 & 593 \\
\bottomrule
\end{tabular}
\caption{Per-model resource accounting.  API total combines OpenAI (USD 161), Gemini (USD 387), and OpenRouter-served open-weight models (USD 46).}
\label{tab:resource-full}
\end{table*}

\section{Efficiency across CogniLoad dimensions}
\label{app:difficulty}

\begin{figure*}[t]
\centering
\includegraphics[width=\textwidth]{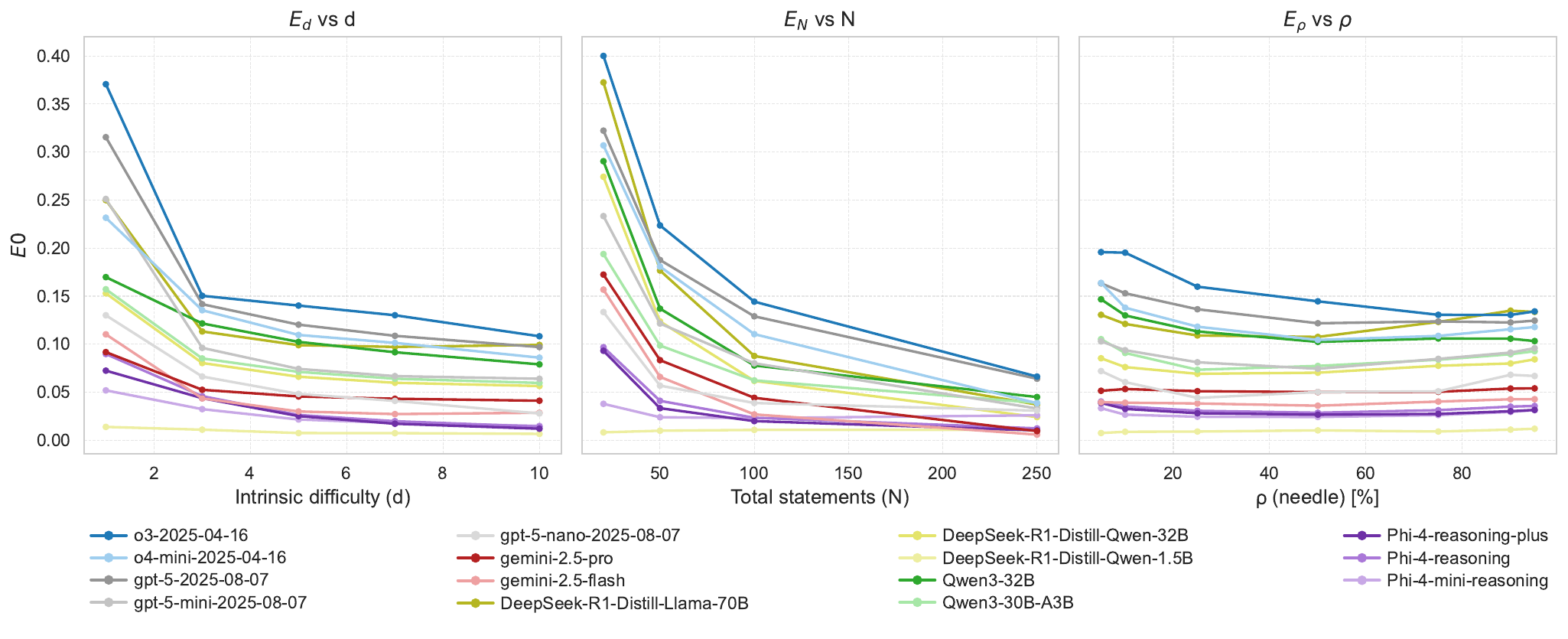}
\caption{Token efficiency $E_0$ across CogniLoad dimensions.  Task length $N$ is the dominant bottleneck: efficiency drops roughly 70--90\% from $N=20$ to $N=250$.  Intrinsic difficulty shows diminishing marginal effects after the lowest levels, while needle fraction has weaker and often U-shaped effects.}
\label{fig:efficiency-by-param-appendix}
\end{figure*}

\paragraph{Task length.}
Task length is the strongest driver of efficiency degradation.  As $N$ increases from 20 to 250 statements, models must process longer sequential dependencies and also generate longer explanations.  The efficiency loss combines rising token use with declining success probability.

\paragraph{Intrinsic difficulty.}
Intrinsic difficulty controls the number of people and attributes as well as the distribution of condition and update clauses.  The observed pattern is a steep drop from trivial to non-trivial settings followed by diminishing marginal effects.  This suggests that many models adopt a broadly similar reasoning strategy once the problem is complex enough to require explicit tracking.

\paragraph{Needle fraction.}
Needle fraction controls the proportion of relevant statements among distractors.  Low needle fractions require sparse relevance detection; high needle fractions require processing a large volume of relevant updates.  The intermediate regime is often most efficient.  This is why the curve is weaker and often U-shaped rather than monotone.

\section{Model-family bottleneck profiles}
\label{app:profiles}

\paragraph{Logic-limited models.}
Small and distilled models often lose efficiency through $r_{\mathrm{logic}}$.  MiMo-7B-RL has the lowest overhead ($\bar{VO}=13.7$) but only 6\% accuracy, making its main limitation reasoning fidelity.  DeepSeek-R1-Distill-Qwen-1.5B combines low logic robustness with very high overhead, making it both logic-limited and verbosity-limited.  Qwen3-1.7B and Phi-4-mini-reasoning similarly show large logic deficits relative to o3.

\paragraph{Verbosity-limited models.}
EXAONE-Deep-32B uses roughly $3.5\times$ more tokens per native work unit than o3 while also having a moderate logic deficit.  Phi-4-reasoning-plus has substantial overhead and some context limitation.  These models are plausible candidates for length control or concise distillation, provided accuracy is monitored.

\paragraph{Context-limited models.}
Gemini~2.5~Pro and Gemini~2.5~Flash have high conditional correctness when they complete, but low $r_{\mathrm{ctx}}$ due to budget truncation.  Their efficiency losses are therefore not primarily ordinary reasoning errors.  Budget-aware decoding, chunking, or loop detection is better aligned with their bottleneck profile than uniform output shortening.

\section{Detailed trace case studies}
\label{app:case-studies}
The following examples illustrate what the compression trace layer adds beyond the outcome and workload layers.  The diagnostic is useful for redundancy and density, but it is not a semantic proof checker.  Excerpts are shortened to protect readability.

\subsection{Degeneracy-dominated collapse: DeepSeek-R1-Distill-Qwen-1.5B}

Summary statistics: $100E_0=1.26$, $r_{\mathrm{ctx}}=0.94$, $r_{\mathrm{logic}}=0.20$, $q_{\mathrm{trace}}=0.052$, $\bar{VO}_{\mathrm{sig}}=7.00$, and $\kappa_{\mathrm{sig}}=0.22$.  Relative to DeepSeek-R1-Distill-Llama-70B, the compression trace decomposition is
\begin{equation}
\Delta\log E_0=-0.06-1.08-1.26-0.55+0.65=-2.30.
\end{equation}
The low $q_{\mathrm{trace}}$ term shows that much of the visible reasoning is highly compressible.  The favorable coupling term does not rescue the poor logic rate and low compressed-signal density.

\begin{lstlisting}
Alright, I need to figure out what Veronica most recently ate based on the given puzzle. Let's break down the information step by step.

First, I'll look at the first sentence: "Veronica is wearing a black hat and is wearing a purple pant and is wearing blue underwear." This tells me that Veronica is wearing a black hat, a purple pant, and blue underwear.

Next, I'll check the subsequent sentences to see if any of the food items mentioned are related to her meal. The next sentence: "The people wearing a purple hat and wearing black pant and having black hair and wearing green socks and wearing yellow underwear and wearing gray gloves and having a blue hat and wearing gray socks and wearing white underwear and having a yellow hat and wearing gray gloves and wearing last ate a sushi and put on gray gloves and wearing white underwear and wearing black socks and change into a green hat and change into a pink hat and change into a green hat and eat a taco ..."

Then, the next sentence: "The people wearing green socks and wearing a red hat and wearing black hat and having green hair and wearing a white hat and wearing green gloves and wearing green hair and wearing a pink hat and wearing green socks and wearing a red hat and wearing a green hat and wearing a green hat and wearing a green hat ..."
\end{lstlisting}

\subsection{Dense but costly verbosity: EXAONE-Deep-32B}

Summary statistics: $100E_0=3.14$, $r_{\mathrm{ctx}}=1.00$, $r_{\mathrm{logic}}=0.35$, $q_{\mathrm{trace}}=0.214$, $\bar{VO}_{\mathrm{sig}}=15.35$, and $\kappa_{\mathrm{sig}}=0.31$.  The model has moderate compression trace quality, but uses much more compressed-signal text per unit of work than the reference.  This is a verbosity-limited rather than a purely redundant profile.

\begin{lstlisting}
Wait, perhaps the exact text may have different conditions. Alternatively, maybe I need to check all statements again.

Statement 30: "The people located in the castle and wearing an orange pant and wearing pink gloves and which most recently read a sci-fi book and which most recently watched a mystery movie and having green hair and wearing a white shirt and wearing a white hat put on white gloves and color their hair to black..."

No.

Alternatively, statement 40: "The people which most recently watched a thriller movie and located in the inn and wearing white gloves and which most recently read a thriller book and wearing a yellow pant and wearing orange underwear change into a white pant and change into a purple shirt and move to the pizzeria..."

Not relevant.

Hmm. Maybe there's no further movement for Anna. Therefore, after statement 2, she's at the port.

Wait, but let me check if there's any other statement that might have moved her again.
\end{lstlisting}

\subsection{Context-limited redundant long reasoning: Phi-4-reasoning-plus}

Summary statistics: $100E_0=3.08$, $r_{\mathrm{ctx}}=0.79$, $r_{\mathrm{logic}}=0.58$, $q_{\mathrm{trace}}=0.149$, $\bar{VO}_{\mathrm{sig}}=11.03$, and $\kappa_{\mathrm{sig}}=0.41$.  The model has both context limitation and low compression density, while still spending high compressed-signal overhead.  The example shows repeated state checks that consume budget, but the metric itself is based only on compression.
\begin{lstlisting}
Rule 58:
"58. The people wearing yellow underwear and which most recently watched an adventure movie change into a gray shirt and put on blue underwear and read a sci-fi book and change into a purple pant and change into yellow socks and listen to pop music."

For Alexander: yellow underwear? yes, underwear is yellow? Wait, check: After rule 8, underwear became gray, then rule 9 didn't change underwear, then rule 10 didn't change underwear, rule 11 didn't, rule 12 didn't, rule 13 didn't, rule 14 didn't, rule 15 didn't, rule 16 didn't, rule 17 didn't, rule 18 didn't, rule 19 didn't, rule 20 didn't, rule 21 didn't, rule 22 didn't, rule 23 didn't, rule 24 didn't, rule 25 didn't, rule 26 didn't, rule 27 didn't, rule 28 didn't, rule 29 didn't, rule 30 didn't, rule 31 didn't, rule 32 didn't, rule 33 didn't, rule 34 didn't, rule 35 didn't, rule 36 didn't, rule 37 didn't, rule 38 didn't, rule 39 didn't, rule 40 didn't, rule 41 didn't, rule 42 didn't, rule 43 didn't, rule 44 didn't, rule 45 didn't, rule 46 didn't, rule 47 didn't, rule 48 didn't, rule 49 didn't, rule 50 didn't, rule 51 didn't, rule 52 didn't, rule 53 didn't, rule 54 didn't, rule 55 didn't, rule 56 didn't.

So currently, after rule 7, underwear became yellow, then rule 8 changed it to gray, then rule 9 didn't change it, so current underwear is gray. So condition "wearing yellow underwear" is false for Alexander. So rule 58 does not apply.
\end{lstlisting}

\end{document}